\documentclass{article}

\usepackage[final]{neurips_2026}

\usepackage{multirow}
\usepackage{graphicx}
\usepackage{float}
\usepackage{longtable}
\usepackage{booktabs}
\usepackage{array}

\usepackage[utf8]{inputenc}
\usepackage[T1]{fontenc}
\usepackage{hyperref}
\usepackage{url}
\usepackage{amsfonts}
\usepackage{amsmath}
\usepackage{nicefrac}
\usepackage{microtype}
\usepackage{xcolor}

\title{Support-Set Target Leakage in Relational Foundation Models during In-Context Learning: Model Dependence and Evaluation Reliability}

\workshoptitle{TAE (Trust-AI-Eval): Can We Trust AI Evaluation?}

\author{%
Roshan Reddy Upendra \\
SAP \\
Palo Alto, CA, USA \\
\texttt{roshan.reddy.upendra@sap.com}\And
Alexandre Dorais \\
SAP \\
Montreal, Canada \\
\texttt{alexandre.dorais@sap.com}\And
Joe Meyer \\
SAP \\
Palo Alto, CA, USA \\
\texttt{joseph.meyer@sap.com}\And
Andrew Pouret \\
SAP \\
Montreal, Canada \\
\texttt{andrew.pouret@sap.com}\And
Anastasios Lambrianos Stappas \\
SAP \\
Montreal, Canada \\
\texttt{anastasios.lambrianos.stappas@sap.com}\And
Dinesh Katupputhur Ramprasath \\
SAP \\
Palo Alto, CA, USA \\
\texttt{dinesh.katupputhur.ramprasath@sap.com}\And
Viswanath Ganapathy \\
SAP \\
\texttt{viswa.ganapathy@sap.com}\And
Tom Palczewski \\
SAP \\
Palo Alto, CA, USA \\
\texttt{tom.palczewski@sap.com}\And
Minghua Li \\
SAP \\
Seattle, WA, USA \\
\texttt{mark.li01@sap.com}
}

\begin{document}

\maketitle

\begin{abstract}

Relational in-context learning (ICL) uses labeled support examples and their linked relational context to predict labels for new queries. This creates a failure mode when target-derived features are present in the support context but unavailable for the query. We study this setting as support-set target leakage. We construct 20 controlled target-derived features that vary in signal fidelity, representation, semantic transparency, coverage, and zero-, one-, and two-hop relational placement, and evaluate them across 13 RelBench tasks and five relational ICL configurations that vary the ICL head, message-passing depth, pretraining cohort, or relational encoder architecture. We evaluate matched 0-hop, 1-hop, and 2-hop leakage settings, together with a Full leakage condition containing all 20 leaker columns. Within the tested configurations, target-table (0-hop) and Full leakage produce the largest aggregate deviations from clean evaluation, while higher-hop effects are often weaker, consistent with differences in effective exposure associated with temporal reachability, sampling, and aggregation fidelity. Leakage effects are strongly task- and model-dependent and can reverse relative conclusions between model variants even when aggregate changes are small. For leaker detection, we compare an Integrated Gradients (IG)-based screening method with mutual information (MI) and leave-one-column-out (LOCO) on a common Baseline subset. Ranking quality is strongest in the high-impact 0-hop and Full leakage conditions, but detector-based removal does not consistently restore the clean evaluation. A four-task rel-salt case study further shows the same evaluation concern with native-schema leakage candidates from the original relational schema. These results identify the support/query information boundary as an important component of reliable relational ICL evaluation.

\end{abstract}

\section{Introduction}

Relational foundation models (RFMs) aim to transfer predictive capabilities across databases, schemas, and tasks. Recent approaches include graph-centric relational encoders such as Griffin \cite{wang2025griffin}, Relational Transformer \cite{ranjan2025relational}, and relational in-context learning (ICL) models that condition predictions on labeled examples from a new task without downstream parameter updates \cite{meyer2026relational}. RDB-PFN \cite{wang2026relational} learns relational ICL from synthetic structural priors, while OpenRFM \cite{chen2026openrfm} studies how relational context, architecture, and pretraining priors affect in-context prediction. Recent systems such as KumoRFM-2 \cite{hudovernik2026kumorfm} similarly support in-context prediction directly over relational data.

Relational ICL introduces an evaluation problem that is less visible in conventional single-table prediction. The support examples are labeled and may be accompanied by records from several linked tables. A column in this relational context may contain information derived from the target or from events occurring after the prediction point. If such information is present in the labeled support context but unavailable when a clean query is evaluated, the model is conditioned on a predictive shortcut that cannot be reproduced for the query. \cite{kaufman2012leakage} describe a related real-world failure in customer prediction, where archived company websites contained references to products that were purchased only later. These references were predictive in historical data but unavailable for new customers, and could not be removed by temporal filtering when earlier website versions were not available. A similar mismatch can occur in relational ICL when target-derived information is present in the labeled support context but unavailable when a clean query is evaluated.

We call this setting support-set target leakage. It differs from leakage caused during dataset construction, temporal splitting, or representation pretraining. The mismatch is introduced during downstream ICL inference: target-derived columns are made available to the support examples while the query schema remains clean.

We study support-set target leakage using 20 synthetic leaker columns varying in signal fidelity, coverage, modality, semantic transparency, and zero-, one-, and two-hop relational placement. We evaluate the resulting support/query mismatch on 13 RelBench tasks across five relational ICL configurations varying message-passing depth, ICL head, pretraining cohort, or relational encoder architecture. Within the tested configurations, target-table and Full leakage produce the largest aggregate deviations, but their magnitude and direction vary substantially across tasks and models. Higher-hop effects are often weak because effective exposure also depends on temporal reachability, neighborhood sampling, and relational aggregation. Support-set contamination can also change the relative ordering of model variants even when aggregate effects are small, motivating our treatment of leakage as an evaluation-reliability problem rather than only a feature-detection problem.

We compare four model dimensions. First, we compare TabPFN and TabICLv2 heads while keeping the relational representation fixed where possible, to test whether leakage sensitivity changes with the downstream ICL head. Second, we compare encoders with four and eight message-passing blocks to study whether message-passing depth changes the effect of zero-, one-, and two-hop leakers. Third, we compare checkpoints pretrained on PluRel-generated data \cite{kothapalli2026plurel} and the CTU--RelBench \cite{motl2015ctu, robinson2024relbench, gu2026relbench} cohort to test whether leakage behavior differs across pretraining settings. Fourth, we compare encoder variants as an additional architecture-level robustness check. We also study whether suspicious columns can be identified and safely removed, comparing Ridge-based Integrated Gradients (IG) with mutual information (MI) and leave-one-column-out (LOCO) on a computationally matched Baseline subset and evaluating ranking, thresholded detection, and downstream mitigation separately. Finally, four multiclass rel-salt \cite{klein2025salt} tasks provide a non-synthetic, native-schema counterpart in which other target-designated fields are candidate leakers.

The study is organized around four research questions: 
\begin{itemize}

    \item \textbf{Impact and model dependence.} How does support-set target leakage affect relational ICL across ICL heads, message-passing depths, pretraining cohorts, and relational encoder families?
    
    \item \textbf{Evaluation reliability.} Are leakage effects stable across datasets and tasks, and can the presence of support-set leakage change relative conclusions between model variants?
    
    \item \textbf{Detection.} How reliably can suspicious relational columns be ranked and detected, how sensitive is thresholded identification to the operating rule, and how does the IG-based detector compare with simpler association-based and direct perturbation baselines?
    
    \item \textbf{Mitigation.}  When detector-flagged columns are removed without using their ground-truth identities, how reliably does the resulting evaluation move toward the clean result, and how do residual leakers and false-positive feature removal affect recovery?

\end{itemize}

Our contributions are:
\begin{enumerate}

    \item We formulate and evaluate support-set target leakage in relational ICL using controlled target-derived columns spanning different leaker types, modalities, semantic transparency, and relational distances, and study how its effects vary across ICL heads, message-passing depths, pretraining cohorts, and encoder families. 
    
    \item We characterize the resulting task-, dataset-, model-, and relational-exposure dependence and show that support-set contamination can reverse conclusions from model comparisons.
    
    \item We compare attribution-, association-, and perturbation-based leaker ranking on a common Baseline subset, evaluate threshold sensitivity, and test detector-based removal as an end-to-end intervention, separating ranking suspicious columns from thresholded identification and safe restoration of the clean evaluation.

\end{enumerate}

\section{Related Work}

Target leakage is a known source of invalid evaluation when training data contain information that would not be available at prediction time. In relational data, such information may also occur in linked tables. 

Predictive Query Language (PQL) \cite{kocijan2026predictive} addresses leakage during relational training-table construction by identifying records that may leak information because they contribute to the target or occur after the prediction anchor time. \cite{xu2026parameter} study observable labels as additional features in relational neighborhoods, while restricting these labels to entities observed before the prediction time to avoid temporal leakage. Our setting differs from both: we study target-derived information that is available in the labeled ICL support context but unavailable to the clean query. Task Scarcity and Label Leakage \cite{azevedo2026task} studies task-specific shortcuts learned during relational transfer, whereas our setting introduces target-derived columns only during downstream ICL inference and leaves model parameters unchanged. 

OpenRFM \cite{chen2026openrfm} perturbs labels and relational features to study whether relational ICL uses its context, and ICLShield \cite{ren2025iclshield} studies poisoned demonstrations in language-model ICL. Recent TabPFN \cite{hu2026noise} work also evaluates robustness to irrelevant and correlated features and label noise. \cite{bilovs2026mechanistic} show that TabPFNv2 \cite{hollmann2025accurate} and TabICLv2 \cite{qu2026tabiclv2} use qualitatively different in-context readout mechanisms and exhibit different failure behavior under targeted perturbations. This motivates testing whether the same support-set leakage affects different ICL heads in the same way. These settings differ from ours, where a feature can be strongly target-predictive in the labeled support examples but unavailable to the clean query.

For detection, we use IG \cite{sundararajan2017axiomatic} over the relational encoder with a Ridge surrogate. GRAFT \cite{sahoo2026graft} similarly uses IG to audit feature reliance in graph neural networks. Association measures such as correlation and MI can identify target-predictive columns but do not show whether the downstream model uses them. LOCO tests the downstream model more directly, but requires repeated evaluations and can be affected by redundant features \cite{breiman2001random, strobl2008conditional}. SHAP-based methods account for feature subsets more broadly, but their cost grows with the number of evaluations and their estimates can depend on how absent or dependent features are handled \cite{lundberg2017unified, aas2021explaining}. Attention scores and intermediate representations provide another view of how information is processed, but they are not by themselves evidence that a feature drives the final output \cite{jain2019attention,wiegreffe2019attention, ravichander2021probing}. VIP-COP \cite{chen2026vip} also selects samples and features from the inference-time context of frozen tabular foundation models to reduce the effect of noisy or irrelevant context, although it does not target support-set leakage. We therefore use IG to screen candidate columns jointly and evaluate attribution ranking, thresholded detection, and downstream removal separately.

\section{Methodology}

\subsection{Problem Formulation}

Let $D$ denote a relational database and $G(D)=(V,E)$ its heterogeneous graph representation. For a downstream task, the model receives labeled support $S=\{(x_i,y_i)\}_{i=1}^{n}$ and predicts labels for query set $Q=\{x_j\}_{j=1}^{m}$, where each $x_i$ contains a target row and the relational context available to the encoder. For a leaker column $c$ stored directly on target row $i$, $z_{i,c}=g_c(y_i)$, where $g_c$ may produce a deterministic, noisy, partially observed, numerical, categorical, or text-valued proxy; higher-hop leakers place related target-derived information on rows connected through relational edges.

Let $S_{\mathrm{clean}}$ denote the original support data and $S_{\mathrm{leaky}}$ the same support after introducing the selected target-derived columns. Our primary comparison for model variant $M$ is $f_M(S_{\mathrm{clean}},Q_{\mathrm{clean}})$ versus $f_M(S_{\mathrm{leaky}},Q_{\mathrm{clean}})$, isolating information available while inferring the task from support but absent when predicting the query. We additionally evaluate $f_M(S_{\mathrm{leaky}},Q_{\mathrm{leaky}})$ as a diagnostic control, where the corresponding injected features are available to both support and query, to distinguish a support--query availability mismatch from features the model does not use.

We quantify the signed leakage effect as
\begin{equation}
\Delta\mathrm{F1}_M =
\mathrm{F1}_M(S_{\mathrm{leaky}},Q_{\mathrm{clean}})
-
\mathrm{F1}_M(S_{\mathrm{clean}},Q_{\mathrm{clean}}),
\end{equation}
and use $|\Delta\mathrm{F1}|$ to summarize deviation magnitude, avoiding either performance inflation or degradation as the defining direction of leakage. An overview of the evaluation and detection--mitigation protocol is provided in Appendix \ref{app:protocol} Figure~\ref{fig:protocol_overview}.

\subsection{Synthetic Leaker Conditions}

We construct a 20-column stress test varying signal fidelity, coverage, representation type, semantic transparency, and relational placement: 14 leakers are placed on the target table, three on a one-hop table, and three on a two-hop table. We also evaluate matched 0-hop-only, 1-hop-only, and 2-hop-only settings using the same three leaker designs at each distance; full construction details are in Appendix \ref{app:leakers}. Only columns assigned to a condition are added to its support schema, and in the support-only setting all injected columns, including their values and feature-name representations, are physically removed from the query schema. Ground-truth leaker identities are used only for construction and detection metrics, not by the detector or to determine removal.

\subsection{Model Comparisons}

We evaluate five relational ICL configurations, changing one component relative to the Baseline while holding the remaining choices fixed. The Baseline uses a four-block Griffin encoder pretrained on the CTU--RelBench cohort with a TabPFN ICL head. MP-8 increases Griffin message-passing depth to eight blocks; PluRel replaces the encoder checkpoint with one pretrained on PluRel-generated relational data; TabICL replaces TabPFN with TabICLv2; and Bimodal replaces Griffin with the Bimodal relational encoder while retaining CTU--RelBench pretraining and TabPFN. Full checkpoint, pretraining-cohort, and architecture details are provided in Appendix~\ref{app:models_info}. 

To evaluate whether leakage changes model-selection conclusions, we compare each variant $M$ with the Baseline $B$. A strict ranking reversal occurs when
\begin{equation}
\mathrm{sign}
\left(
\mathrm{F1}^{\mathrm{clean}}_M-
\mathrm{F1}^{\mathrm{clean}}_B
\right)
\neq
\mathrm{sign}
\left(
\mathrm{F1}^{\mathrm{leaky}}_M-
\mathrm{F1}^{\mathrm{leaky}}_B
\right),
\end{equation}
excluding comparisons tied under either condition. In the primary analysis, variant--Baseline differences with absolute magnitude at or below $10^{-6}$ are treated as ties. We report both the number of reversals and the number of non-tied comparisons, and evaluate sensitivity to larger minimum performance margins in Appendix~\ref{app:full_results}.

The four variants probe distinct components that may influence how a support-side shortcut affects downstream ICL. MP-8 tests whether increasing relational message-passing depth changes sensitivity to leakers placed at different relational distances. PluRel tests whether the encoder pretraining cohort changes leakage sensitivity while retaining the same Griffin architecture and TabPFN head. TabICL isolates the downstream ICL head, testing whether the same relational representation produces a different response when task inference is performed by TabICLv2 rather than TabPFN. Bimodal provides an architecture-level comparison by replacing Griffin with a different relational encoder while retaining the CTU--RelBench pretraining setting and TabPFN head.

\subsection{Attribution-Based Detection}

We use IG \cite{sundararajan2017axiomatic} as an attribution-based screening method for candidate relational columns. We extract the leaky training representations and fit a Ridge classifier as a differentiable surrogate, using an internal 80/20 split to check surrogate fit with stratification when class counts permit. The fitted Ridge classifier is evaluated on task validation representations, and IG is computed through the frozen relational encoder on validation batches.

For each attributed column, its encoded representation is interpolated from a zero-embedding baseline to the observed representation using 12 midpoint integration steps, resetting the relational sampling seed at each point so that the sampled neighborhood and internal few-shot context remain fixed. The objective is the margin between the Ridge-predicted class under the full input and its strongest alternative; column attributions combine the overall mean and the mean among the strongest 10\% of occurrences by geometric mean. These scores measure influence under the Ridge surrogate rather than exact attribution of the downstream ICL prediction. The actual ICL configurations therefore measure leakage impact and post-removal performance, while Ridge-IG provides the screening signal.

\subsection{Detector Baselines and Thresholding}

We compare Ridge-IG with MI and LOCO. MI is a model-independent association baseline that ranks columns by dependence on support labels without using the relational encoder or attribution surrogate. LOCO is a model-based perturbation baseline that measures the downstream effect of removing each candidate column and requires no differentiable attribution path. MI does not establish downstream reliance, whereas LOCO more directly measures it but requires a separate evaluation per candidate and can underestimate importance under redundant features. We primarily evaluate continuous rankings, including AUPRC, before applying a decision threshold.

For IG-based removal, aggregated scores are log-transformed and robustly standardized within source-table and feature-family groups using the median and scaled median absolute deviation, with global statistics for groups containing fewer than ten scored columns. Columns above the empirical 90th percentile of robust z-scores in a run are flagged. This q90 rule restricts removal to the highest-scoring candidates without assuming a calibrated null distribution and is a heuristic rather than a statistically principled threshold; ground-truth leaker identities are not used. Appendix \ref{app:detection} additionally evaluates q80, q95, and Bonferroni-style and BH-style operating rules.

\subsection{Mitigation}

For mitigation, every detector-flagged column, including false-positive original columns, is removed from both support and query schemas; relational embeddings are then recomputed and ICL inference repeated. Because leakage can increase or decrease macro-F1, we evaluate mitigation by distance from the clean result and define recovery gain as
\begin{equation}
G =
\left|
\mathrm{F1}_{\mathrm{leaky}}
-
\mathrm{F1}_{\mathrm{clean}}
\right|
-
\left|
\mathrm{F1}_{\mathrm{post}}
-
\mathrm{F1}_{\mathrm{clean}}
\right|.
\end{equation}
Thus $G>0$ means removal moves the evaluation closer to clean. Recovery can fail because active leakers remain or useful original columns are removed as false positives. We also summarize Baseline IG leaker-family detection and associated recovery; because several families may be removed in one run, these family-level quantities are descriptive associations rather than causal effects.

\subsection{Experimental Setup}

\subsubsection{Primary Tasks}

We evaluate the main experiments on 13 binary classification tasks from RelBench databases \cite{robinson2024relbench, gu2026relbench}. The benchmark includes tasks from \texttt{rel-avito}, \texttt{rel-event}, \texttt{rel-trial}, and \texttt{rel-arxiv}. All tasks use their standard train, validation, and test splits. For each task, the training and validation target-row embeddings form the labeled ICL support set, while the test embeddings form the query set. The same leakage construction and support/query conditions are used across the primary tasks. The evaluation conditions and controlled leaker design are summarized in Appendix Figure~\ref{fig:figure1_protocol_design_validity}. 

\subsubsection{Native-Schema Target-Field Leakage Case Study}

We additionally evaluate rel-salt as a separate four-task multiclass case study \cite{klein2025salt}. Unlike the controlled RelBench experiments, we do not inject synthetic target-derived columns. Instead, for each prediction task we treat the remaining SALT target-designated fields as candidate leakers when they are included in the relational context. For example, when predicting \texttt{item-incoterms}, other prediction targets such as \texttt{sales-incoterms}, \texttt{item-plant}, and \texttt{sales-office} may provide information outside the intended feature set for that prediction setting. Using the Baseline model, we compare performance with and without these fields and apply the same detector-based removal and downstream recovery criterion. \texttt{rel-salt} is reported separately and excluded from cross-model aggregates; per-task results are in Appendix \ref{app:salt}. 

\section{Results and Discussion}

Figure \ref{fig:figure1_protocol_design_validity} confirms that the injected leakers contain usable target information: when corresponding features are available to both support and query, the matched-availability control produces large gains for the 0-hop and Full conditions, whereas restricting the same information to support can instead reduce performance. We use the matched condition only as an exploitability check; the failure studied here is that the model infers a task from a support representation that cannot be reproduced at query time.

\begin{figure}[t]
\centering
\includegraphics[width=\textwidth]{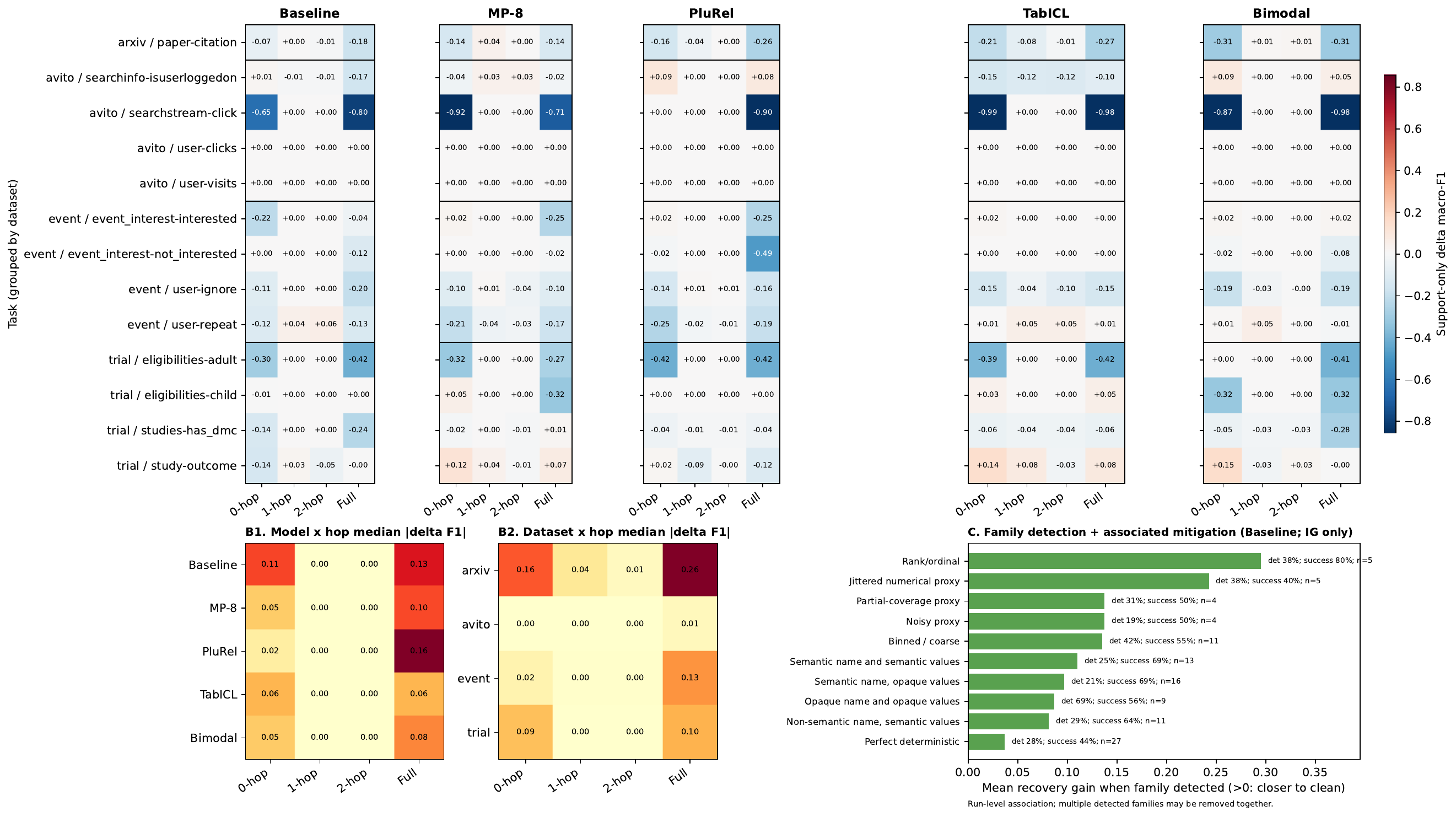} 
\caption{
Support-set leakage is structured by task, model, relational exposure, and leaker type.
(A) Signed macro-F1 change under leaky support/clean query across 13 tasks, five configurations, and 0-hop, 1-hop, 2-hop, and Full leakage.
(B1) Median $|\Delta F1|$ by model and hop; (B2) by dataset and hop.
(C) Baseline IG leaker-family q90 detection and associated recovery: \emph{det} is the fraction of evaluated true-leaker columns flagged; $n$ is the number of runs with at least one family member flagged; and \emph{success} is the fraction of those runs with valid recovery satisfying $|F1_{\mathrm{post}}-F1_{\mathrm{clean}}|+10^{-6}<|F1_{\mathrm{leaky}}-F1_{\mathrm{clean}}|$. Multiple families may be removed together, so these are associations rather than causal family effects. Full task-, leaker-, and detector-level results are provided in Appendix \ref{app:detection}, \ref{app:individual_leakers}, \ref{app:full_results}.
}
\label{fig:figure2_model_task_hop_leaker_type}
\end{figure}

Figure \ref{fig:figure2_model_task_hop_leaker_type}B1--B2 shows that target-table and Full leakage produce the largest aggregate deviations across the five configurations, while nominal hop count alone does not explain higher-hop effects. Median $|\Delta F1|$ at 0-hop ranges from 0.02 to 0.11 across models, compared with medians rounding to zero at 1 and 2 hops; Full leakage ranges from 0.06 to 0.16. Dataset-level aggregation shows the same structure: \texttt{rel-arxiv} is most sensitive, with median $|\Delta F1|=0.16$ at 0-hop and $0.26$ under Full leakage, whereas \texttt{rel-avito} is near zero because several tasks are invariant despite large effects on \texttt{searchstream-click}. Appendix \ref{app:individual_leakers} shows that the target-table effect is not driven by a single synthetic construction.

The weak higher-hop averages do not imply that relational leakage disappears beyond the target table. Effective exposure also depends on temporal reachability, neighborhood sampling, and relational aggregation: selected higher-hop paths are temporally unavailable for several tasks and only partially reachable for others, while two-hop aggregation can turn an exact target copy into a noisier group-level proxy. For example, the study-outcome 2-hop aggregate agrees with the queried label only about 65\% of the time. Nominal hop count therefore does not characterize effective exposure by itself. Selected path-availability and leaker-construction diagnostics are provided in Appendix~\ref{app:leakers}. 

Figure \ref{fig:figure2_model_task_hop_leaker_type}A shows strong task dependence that aggregate statistics can hide. \texttt{searchstream-click} exhibits large support-only degradation, whereas \texttt{user-clicks} and \texttt{user-visits} remain unchanged across all five configurations. The latter are highly imbalanced and their clean predictions are already close to constant-class solutions, limiting changes in macro-F1, although identical aggregate scores need not imply identical predictions. Leakage can also increase macro-F1, as for \texttt{study-outcome} under some 0-hop configurations; such increases do not imply beneficial leakage or improved generalization, because a support-induced change in the learned rule can accidentally improve a particular evaluation set. We therefore measure severity by deviation from the valid clean evaluation rather than assuming a fixed direction.

Identical clean F1 values likewise need not imply identical model behavior. On \texttt{event-not-interested}, for example, the models begin from the same majority-class solution and all exploit matched Full leakage, yet under support-only Full leakage TabICL remains near that solution while PluRel collapses toward the opposite class. This indicates configuration-specific sensitivity to support/query schema mismatch rather than different information content in the dataset.

Furthermore, our results indicate that clean model quality also does not imply robustness to support-set leakage. Message-passing depth, pretraining cohort, ICL head, and encoder architecture change the magnitude and sometimes the sign of the response, but no evaluated configuration eliminates the failure or yields a universal robustness ordering. PluRel has the smallest median 0-hop effect (0.02) but the largest Full effect (0.16), whereas TabICL has a larger 0-hop median (0.06) but the smallest Full median (0.06). The extreme PluRel Full failures occur despite strong matched-availability performance, indicating sensitivity to the missing-query schema rather than inability to exploit target-derived information. Because other variables are held fixed within each comparison, each tested model dimension can alter leakage sensitivity, but no single choice characterizes or eliminates it.

\begin{figure}[t]
\centering
\includegraphics[width=\textwidth]{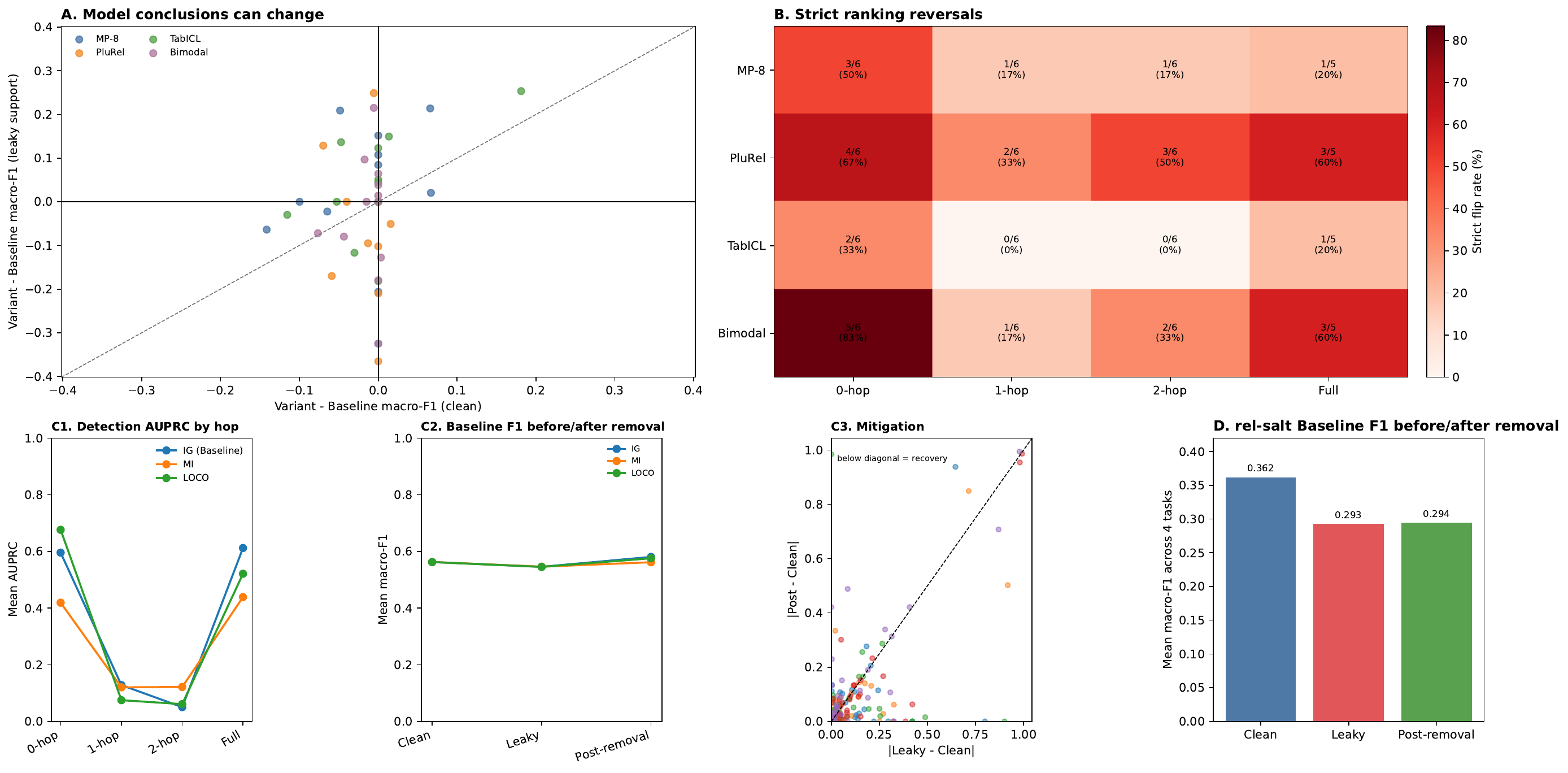}
\caption{
\textbf{Support-set leakage can change model-selection conclusions and is not reliably repaired by automatic column removal.}
\textbf{(A)} Variant-versus-Baseline macro-F1 advantages under clean and support-only leakage; axis crossings indicate changes in model ordering.
\textbf{(B)} Strict ranking reversals among non-tied comparisons, reported as count/denominator and percentage.
\textbf{(C1)} Mean AUPRC for IG, MI, and LOCO on the common Baseline detector subset
\textbf{(C2)} Mean Baseline macro-F1 before leakage, under leakage, and after detector-based removal. MI and LOCO are restricted to this subset because of their computational cost.
\textbf{(C3)} Distance from clean before versus after removal; points below the diagonal move toward the clean evaluation.
\textbf{(D)} \texttt{rel-salt} case study averaged over four Baseline tasks, comparing clean, target-field-contaminated, and post-removal performance.
}
\label{fig:figure3_reliability_detection_mitigation_salt}
\end{figure}

Figure \ref{fig:figure3_reliability_detection_mitigation_salt}A compares each variant's advantage over the Baseline under clean and support-only leakage; an axis crossing changes which model would be preferred. Among non-tied comparisons, 0-hop leakage reverses the ordering in 3/6 MP-8, 4/6 PluRel, 2/6 TabICL, and 5/6 Bimodal comparisons; under Full leakage the counts are 1/5, 3/5, 1/5, and 3/5. Even 1-hop and 2-hop conditions can reverse rankings despite near-zero median $|\Delta F1|$, because small task-specific changes can exceed the often-small clean margin between models even when they cancel in aggregate. We therefore report reversal counts and denominators rather than treating the percentages as population estimates. Importantly, the ranking-reversal result is not driven solely by near-tied model comparisons: 24 of the 32 strict reversals remain when both clean and leaky variant--Baseline margins are required to exceed 0.005, and 20 remain at a 0.01 margin (Appendix~\ref{app:full_results}). These observations reinforce the most consequential evaluation result: leakage can change the conclusion of a model comparison even when its average effect is small.

On the common Baseline subset where all three detectors are feasible, Figure \ref{fig:figure3_reliability_detection_mitigation_salt}C compares Ridge-IG, MI, and LOCO. Ranking quality is generally stronger at 0-hop and Full leakage than at higher hops, consistent with weaker or less available higher-hop target signal and, for IG and LOCO, limited downstream exposure. This pattern persists across q80, q90, q95, and Bonferroni-style and BH-style operating rules. In the high-impact 0-hop and Full conditions, q90 also yields higher detection F1 than the Bonferroni-style and BH-style alternatives; no percentile dominates, with q80 favoring recall and q95 precision, so q90 is treated as an intermediate rather than optimized operating point. Reviewing the top five columns identifies at least one leaker in 88\% of 0-hop runs and 94\% of Full runs, while remaining substantially less effective at higher hops. Full ranking and threshold-sensitivity results are in Appendix~\ref{app:detection}.

Mitigation shows why ranking quality alone is insufficient. Post-removal performance often moves toward clean but not consistently, because detector-based removal creates a fourth input distribution: undetected leakers remain in support, false-positive original columns are removed from both support and query, and some featureless node types receive placeholder representations. For example, TabICL \texttt{user-repeat} removes all active 0-hop leakers but also ten original columns, worsening performance. Conversely, performance above the clean baseline does not establish successful cleaning when residual leakage and original-feature removal remain. 

Additionally, the \texttt{rel-salt} case study tests the same principle with native-schema rather than synthetic leakage candidates (Figure \ref{fig:figure3_reliability_detection_mitigation_salt}D). Across four multiclass Baseline tasks, retaining other target-designated SALT fields reduces mean macro-F1 from 0.362 to 0.293; detector-based removal reaches 0.294 and does not restore clean evaluation. Although limited to one configuration, this shows that the support/query information-boundary problem is not restricted to engineered proxies: fields that are valid prediction targets can still be invalid inputs for another prediction-time setting. Per-task results are in Appendix \ref{app:salt}.

\section{Conclusion}

We studied support-set target leakage as an inference-time evaluation failure in relational ICL, where target-derived information is available while the model infers a task from labeled support examples but unavailable to the evaluated query. Across 13 RelBench tasks, five relational ICL configurations, controlled leakers spanning multiple representations and relational distances, and a native-schema rel-salt case study, three conclusions emerge. First, leakage severity is not determined by column count or nominal hop number alone: higher-hop effects depend on effective relational exposure, including temporal reachability, sampling, and aggregation fidelity. Second, leakage is highly task- and model-dependent and can reverse model rankings, so support-only leakage can invalidate model-selection conclusions even when aggregate changes are small. Third, detection and remediation are distinct: the stronger 0-hop/Full detection pattern persists across threshold choices, while useful rankings do not guarantee restoration of clean evaluation because residual leakage and false-positive removal create a new input configuration.

To our knowledge, this is the first systematic evaluation of the support-only target-derived feature availability mismatch in relational ICL, including its consequences for model comparison, detection, and mitigation. The practical implication is that the support/query information boundary should be part of the evaluation protocol: benchmarks should audit whether support features are available under intended prediction-time semantics, examine robustness across support conditions rather than a single score, and validate mitigation by downstream recovery rather than detector quality alone.

\bibliographystyle{unsrtnat}
\bibliography{reference}

@article{ranjan2025relational,
  title={Relational Transformer: Toward Zero-Shot Foundation Models for Relational Data},
  author={Ranjan, Rishabh and Hudovernik, Valter and Znidar, Mark and Kanatsoulis, Charilaos and Upendra, Roshan and Mohammadi, Mahmoud and Meyer, Joe and Palczewski, Tom and Guestrin, Carlos and Leskovec, Jure},
  journal={arXiv preprint arXiv:2510.06377},
  year={2025}
}

@article{wang2025griffin,
  title={Griffin: Towards a graph-centric relational database foundation model},
  author={Wang, Yanbo and Wang, Xiyuan and Gan, Quan and Wang, Minjie and Yang, Qibin and Wipf, David and Zhang, Muhan},
  journal={arXiv preprint arXiv:2505.05568},
  year={2025}
}

@article{hudovernik2026kumorfm,
  title={KumoRFM-2: Scaling foundation models for relational learning},
  author={Hudovernik, Valter and L{\'o}pez, Federico and Kocijan, Vid and Nitta, Akihiro and Lenssen, Jan Eric and Leskovec, Jure and Fey, Matthias},
  journal={arXiv preprint arXiv:2604.12596},
  year={2026}
}

@article{wang2026relational,
  title={Relational In-Context Learning via Synthetic Pre-training with Structural Prior},
  author={Wang, Yanbo and You, Jiaxuan and Shi, Chuan and Zhang, Muhan},
  journal={arXiv preprint arXiv:2603.03805},
  year={2026}
}

@article{chen2026openrfm,
  title={OpenRFM: Dissecting Relational In-Context Learning},
  author={Chen, Zhikai and Yin, Junyu and Gu, Jialiang and Xiong, Siheng and Liu, Xiaoze and Zhang, Ruowang and Zhou, Keren and Guo, Kai},
  journal={arXiv preprint arXiv:2606.04320},
  year={2026}
}

@article{kaufman2012leakage,
  title={Leakage in data mining: Formulation, detection, and avoidance},
  author={Kaufman, Shachar and Rosset, Saharon and Perlich, Claudia and Stitelman, Ori},
  journal={ACM Transactions on Knowledge Discovery from Data (TKDD)},
  volume={6},
  number={4},
  pages={1--21},
  year={2012},
  publisher={ACM New York, NY, USA}
}

@article{ren2025iclshield,
  title={ICLShield: Exploring and Mitigating In-Context Learning Backdoor Attacks},
  author={Ren, Zhiyao and Liang, Siyuan and Liu, Aishan and Tao, Dacheng},
  journal={arXiv preprint arXiv:2507.01321},
  year={2025}
}

@inproceedings{sundararajan2017axiomatic,
  title={Axiomatic attribution for deep networks},
  author={Sundararajan, Mukund and Taly, Ankur and Yan, Qiqi},
  booktitle={International conference on machine learning},
  pages={3319--3328},
  year={2017},
  organization={PMLR}
}

@article{robinson2024relbench,
  title={Relbench: A benchmark for deep learning on relational databases},
  author={Robinson, Joshua and Ranjan, Rishabh and Hu, Weihua and Huang, Kexin and Han, Jiaqi and Dobles, Alejandro and Fey, Matthias and Lenssen, Jan E and Yuan, Yiwen and Zhang, Zecheng and others},
  journal={Advances in Neural Information Processing Systems},
  volume={37},
  pages={21330--21341},
  year={2024}
}

@article{motl2015ctu,
  title={The CTU prague relational learning repository},
  author={Motl, Jan and Schulte, Oliver},
  journal={arXiv preprint arXiv:1511.03086},
  year={2015}
}

@article{sahoo2026graft,
  title={GRAFT: Auditing Graph Neural Networks via Global Feature Attribution},
  author={Sahoo, Rishi Raj and Mishra, Subhankar},
  journal={arXiv preprint arXiv:2605.03377},
  year={2026}
}

@article{hu2026noise,
  title={Noise Immunity in In-Context Tabular Learning: An Empirical Robustness Analysis of TabPFN's Attention Mechanisms},
  author={Hu, James and Ghelichi, Mahdi},
  journal={arXiv preprint arXiv:2604.04868},
  year={2026}
}

@article{azevedo2026task,
  title={Task Scarcity and Label Leakage in Relational Transfer Learning},
  author={Azevedo, Francisco Galuppo and Loures, Clarissa Lima and Correa, Denis Oliveira},
  journal={arXiv preprint arXiv:2603.29914},
  year={2026}
}

@inproceedings{meyer2026relational,
  title={Relational in-context learning on structured data via neighborhood aggregation and structural information},
  author={Meyer, Joe and Palczewski, Tom and Shaikh, Afreen and Mohammadi, Mahmoud and Ramprasath, Dinesh Katupputhur and Paresh, Karan and Upendra, Roshan Reddy and Li, Mark},
  booktitle={Proceedings of the AAAI Symposium Series},
  volume={9},
  number={1},
  pages={219--220},
  year={2026}
}

@article{kothapalli2026plurel,
  title={PluRel: synthetic data unlocks scaling laws for relational foundation models},
  author={Kothapalli, Vignesh and Ranjan, Rishabh and Hudovernik, Valter and Dwivedi, Vijay Prakash and Hoffart, Johannes and Guestrin, Carlos and Leskovec, Jure},
  journal={arXiv preprint arXiv:2602.04029},
  year={2026}
}

@article{gu2026relbench,
  title={Relbench v2: A large-scale benchmark and repository for relational data},
  author={Gu, Justin and Ranjan, Rishabh and Kanatsoulis, Charilaos and Tang, Haiming and Jurkovic, Martin and Hudovernik, Valter and Znidar, Mark and Chaturvedi, Pranshu and Shroff, Parth and Li, Fengyu and others},
  journal={arXiv preprint arXiv:2602.12606},
  year={2026}
}

@article{breiman2001random,
  title={Random forests},
  author={Breiman, Leo},
  journal={Machine learning},
  volume={45},
  number={1},
  pages={5--32},
  year={2001},
  publisher={Springer}
}

@article{strobl2008conditional,
  title={Conditional variable importance for random forests},
  author={Strobl, Carolin and Boulesteix, Anne-Laure and Kneib, Thomas and Augustin, Thomas and Zeileis, Achim},
  journal={BMC bioinformatics},
  volume={9},
  number={1},
  pages={307},
  year={2008},
  publisher={Springer}
}

@article{lundberg2017unified,
  title={A unified approach to interpreting model predictions},
  author={Lundberg, Scott M and Lee, Su-In},
  journal={Advances in neural information processing systems},
  volume={30},
  year={2017}
}

@article{aas2021explaining,
  title={Explaining individual predictions when features are dependent: More accurate approximations to Shapley values},
  author={Aas, Kjersti and Jullum, Martin and L{\o}land, Anders},
  journal={Artificial Intelligence},
  volume={298},
  pages={103502},
  year={2021},
  publisher={Elsevier}
}

@inproceedings{jain2019attention,
  title={Attention is not explanation},
  author={Jain, Sarthak and Wallace, Byron C},
  booktitle={Proceedings of the 2019 Conference of the North American Chapter of the Association for Computational Linguistics: Human Language Technologies, Volume 1 (Long and Short Papers)},
  pages={3543--3556},
  year={2019}
}

@inproceedings{wiegreffe2019attention,
  title={Attention is not not explanation},
  author={Wiegreffe, Sarah and Pinter, Yuval},
  booktitle={Proceedings of the 2019 conference on empirical methods in natural language processing and the 9th international joint conference on natural language processing (EMNLP-IJCNLP)},
  pages={11--20},
  year={2019}
}

@inproceedings{ravichander2021probing,
  title={Probing the probing paradigm: Does probing accuracy entail task relevance?},
  author={Ravichander, Abhilasha and Belinkov, Yonatan and Hovy, Eduard},
  booktitle={Proceedings of the 16th Conference of the European Chapter of the Association for Computational Linguistics: Main Volume},
  pages={3363--3377},
  year={2021}
}

@article{kocijan2026predictive,
  title={Predictive Query Language: A Domain-Specific Language for Predictive Modeling on Relational Databases},
  author={Kocijan, Vid and Sunil, Jinu and Lenssen, Jan Eric and Deb, Viman and Xe, Xinwei and Gomez, Federico Reyes and Fey, Matthias and Leskovec, Jure},
  journal={arXiv preprint arXiv:2602.09572},
  year={2026}
}

@article{bilovs2026mechanistic,
  title={A Mechanistic Study of Tabular Foundation Models},
  author={Bilo{\v{s}}, Marin and Wilson, James T and Schneider, Anderson and Nevmyvaka, Yuriy},
  journal={arXiv preprint arXiv:2605.21288},
  year={2026}
}

@article{qu2026tabiclv2,
  title={TabICLv2: A better, faster, scalable, and open tabular foundation model},
    author={Qu, Jingang and Holzm{\"u}ller, David and Varoquaux, Ga{\"e}l and Morvan, Marine Le},
  journal={arXiv preprint arXiv:2602.11139},
  year={2026}
}

@article{hollmann2025accurate,
  title={Accurate predictions on small data with a tabular foundation model},
  author={Hollmann, Noah and M{\"u}ller, Samuel and Purucker, Lennart and Krishnakumar, Arjun and K{\"o}rfer, Max and Hoo, Shi Bin and Schirrmeister, Robin Tibor and Hutter, Frank},
  journal={Nature},
  volume={637},
  number={8045},
  pages={319--326},
  year={2025},
  publisher={Nature Publishing Group UK London}
}

@article{chen2026vip,
  title={VIP-COP: Context Optimization for Tabular Foundation Models},
  author={Chen, Yilong and Ding, Xueying and Akoglu, Leman},
  journal={arXiv preprint arXiv:2605.12904},
  year={2026}
}

@article{klein2025salt,
  title={SALT: Sales autocompletion linked business tables dataset},
  author={Klein, Tassilo and Biehl, Clemens and Costa, Margarida and Sres, Andre and Kolk, Jonas and Hoffart, Johannes},
  journal={arXiv preprint arXiv:2501.03413},
  year={2025}
}

@article{xu2026parameter,
  title={Parameter-Free Encoders Remain Viable for RDB Foundation Models},
  author={Xu, Linjie and Wipf, David},
  journal={arXiv preprint arXiv:2607.05476},
  year={2026}
}

\appendix

\section{Experimental Protocol and Effective Relational Exposure}
\label{app:protocol}

For each RelBench task, the training and validation target-row embeddings form the labeled ICL support set, while the test embeddings form the query set. Each target-row embedding is computed from its relational neighborhood using the same sampling configuration within a model comparison. The downstream TabPFN evaluation fits on the concatenated training and validation embeddings and predicts the test embeddings.

For the clean condition, no injected columns are present. For the primary support-only--leakage condition, the selected leaker columns are injected in the training and validation support data but not injected to the test query schema, including both their values and feature-name representations. For the matched-availability diagnostic, the corresponding injected columns are instead retained in the test query as well. After detector-based mitigation, every flagged column is removed from both support and query schemas, and all relational embeddings are recomputed. The pretrained encoder remains frozen throughout evaluation.

The matched-availability condition is used only as an exploitability control. In this condition, the same selected target-derived features are available to both the labeled support set and the test query. Thus, $Q_{\mathrm{leaky}}$ denotes the test query with the corresponding injected features present. This condition is not used for the primary support-only model comparisons, ranking-reversal analysis, or  mitigation evaluation.

\begin{figure}[t]
\centering
\includegraphics[width=\textwidth]{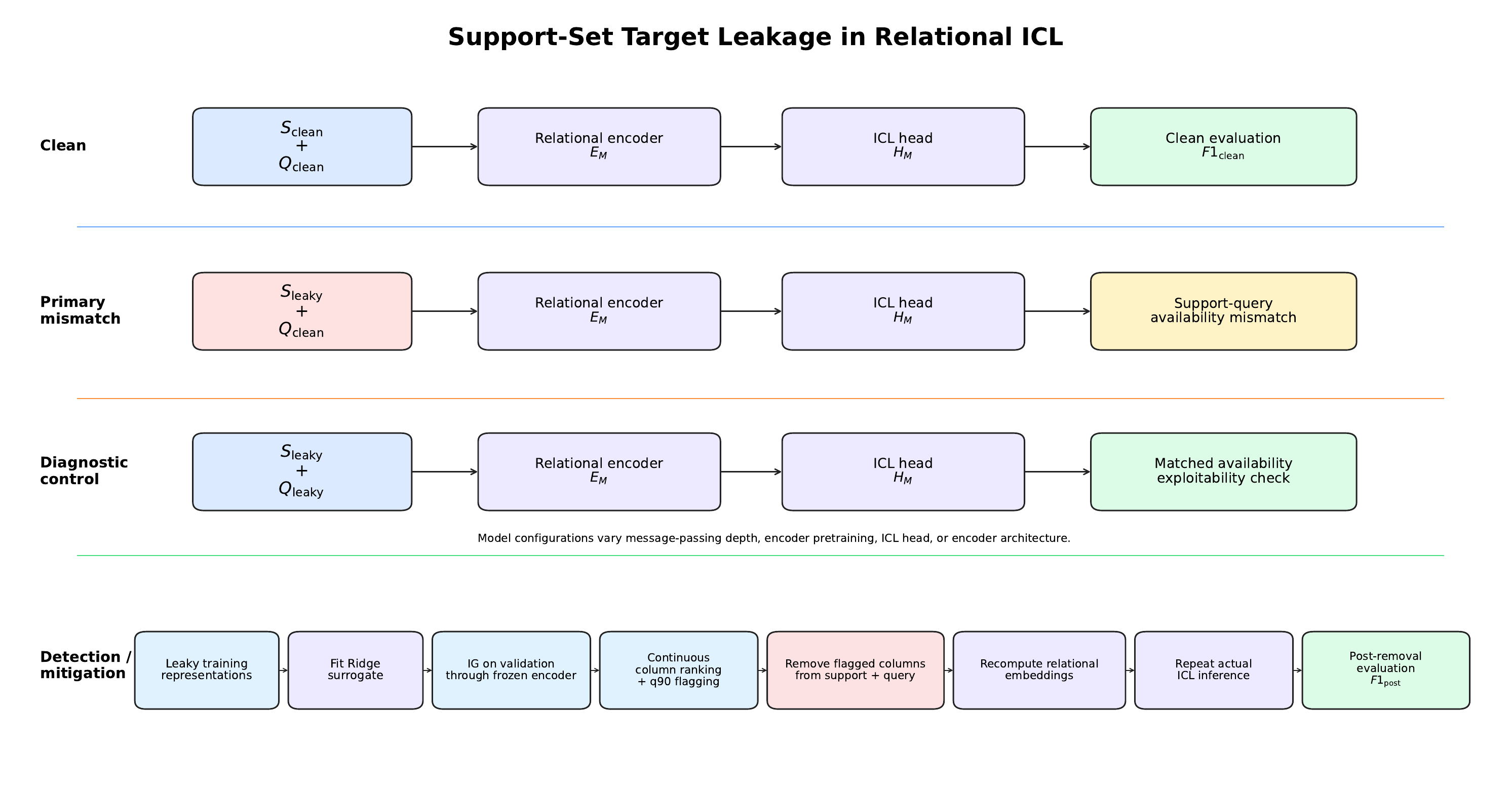} 
\caption{
Overview of the evaluation and detection--mitigation protocol.
The primary support-set leakage condition compares
$(S_{\mathrm{leaky}},Q_{\mathrm{clean}})$ with the clean evaluation
$(S_{\mathrm{clean}},Q_{\mathrm{clean}})$, while
$(S_{\mathrm{leaky}},Q_{\mathrm{leaky}})$ is used only as a
matched-availability diagnostic. For detection, Ridge-IG provides a
screening signal; columns flagged at the q90 operating point are removed
from both support and query, relational embeddings are recomputed, and the
corresponding ICL configuration is reevaluated.
}
\label{fig:protocol_overview}
\end{figure}

\begin{figure}[t]
\centering
\includegraphics[width=1.0\linewidth]{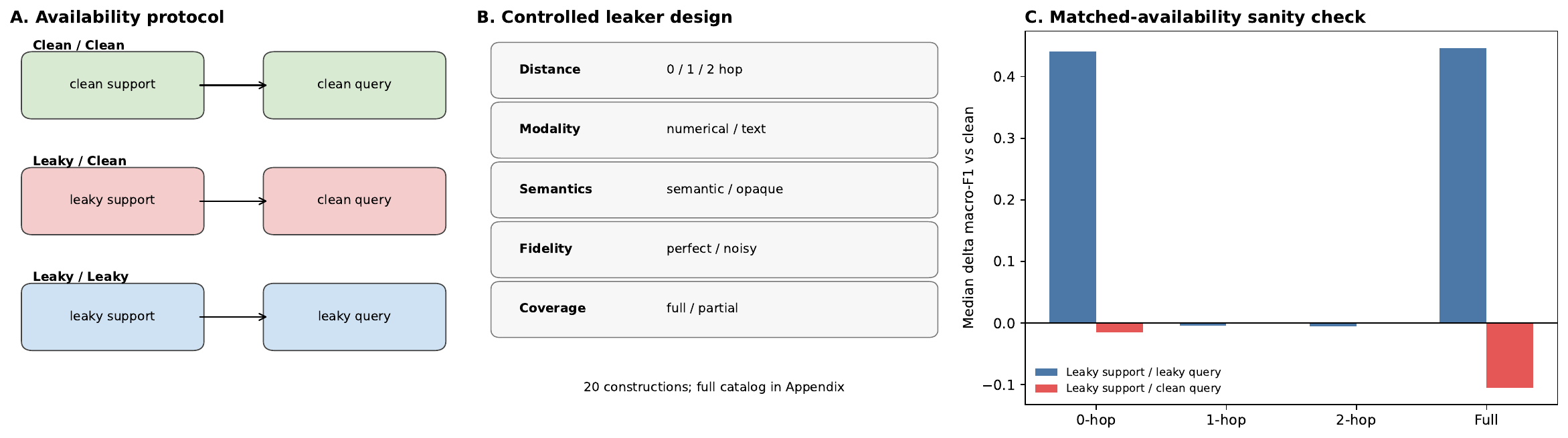} 
\caption{
\textbf{Support-set target leakage protocol and controlled stress test.}
\textbf{(A)} Evaluation conditions: clean support/clean query, leaky support/clean query (the primary availability-mismatch setting), and leaky support/leaky query (diagnostic control).
\textbf{(B)} The 20 synthetic constructions vary relational distance, modality, semantic transparency, signal fidelity, and coverage; the full catalog is provided in Appendix \ref{app:leakers}.
\textbf{(C)} Median macro-F1 change relative to clean under leaky support / clean query and matched-availability (leaky support / leaky query) conditions. 
}
\label{fig:figure1_protocol_design_validity}
\end{figure}

\section{Model Configurations and Pretraining Data}
\label{app:models_info}

\subsection{Model Details}

We evaluate five relational ICL configurations that vary one component of the Baseline pipeline at a time (Table \ref{tab:model_configs}). The Griffin encoder combines attention-based aggregation over encoded row attributes with relation-aware message passing. Within each message-passing block, neighboring representations are mean-aggregated separately by relation type, combined with relation embeddings, and max-aggregated across relations. Gated forward- and reverse-edge messages are added through residual updates. Our Baseline uses four message-passing blocks and produces 512-dimensional target-row representations from two-hop neighborhoods sampled with fanout 9; MP-8 increases the message-passing depth to eight blocks. The Baseline encoder is pretrained on the CTU–RelBench cohort and used with a pretrained TabPFN ICL head. The ICL head conditions on the labeled training and validation representations to predict the test queries. No downstream parameter updates are applied to the relational encoder. All configurations use the same support/query protocol, selected data, and neighborhood-sampling settings within each comparison.

\begin{table}[H]
\centering
\small
\caption{\textbf{Model configurations.} Each variant changes the indicated component relative to the Baseline while retaining the remaining evaluation configuration.}
\label{tab:model_configs}
\begin{tabular}{lccccp{2.6cm}}
\toprule
Setting & Encoder & MP depth & Pretraining & ICL head & Change from Baseline \\
\midrule
Baseline & Griffin & 4 & CTU--RelBench & TabPFN & Baseline reference configuration \\
MP-8     & Griffin & 8 & CTU--RelBench & TabPFN & Message-passing depth \\
PluRel   & Griffin & 4 & PluRel & TabPFN & Pretraining cohort \\
TabICL   & Griffin & 4 & CTU--RelBench & TabICLv2 & ICL head \\
Bimodal  & Bimodal & 4 & CTU--RelBench & TabPFN & Relational encoder \\
\bottomrule
\end{tabular}
\end{table}

The MP-8 configuration increases the Griffin message-passing depth from four to eight message-passing blocks while retaining the Baseline pretraining and ICL head. The PluRel configuration uses the same Griffin architecture and TabPFN head as Baseline but replaces the CTU--RelBench-pretrained encoder with a checkpoint pretrained on PluRel-generated relational data. The TabICL configuration retains the Baseline Griffin representation and replaces TabPFN with the pretrained TabICLv2 ICL head. The Bimodal configuration replaces Griffin with a heterogeneous dual-attention encoder while retaining the four-layer depth, CTU--RelBench pretraining setting, and TabPFN head. Whereas Griffin uses relation-aware neighborhood aggregation, Bimodal combines relation-conditioned attention over local relational neighborhoods with a parallel type-agnostic attention pathway, followed by learned fusion. All configurations use 512-dimensional relational representations and neighborhood fanout 9.

\subsection{Pretraining data}

The CTU--RelBench encoder checkpoints were pretrained on clean relational data (i.e., no synthetic leaker columsn injected) from 15 CTU databases~\cite{motl2015ctu} and four RelBench  databases~\cite{robinson2024relbench}. The complete pretraining set is listed in Table~\ref{tab:pretraining_datasets}. The downstream evaluation databases \texttt{rel-arxiv}, \texttt{rel-avito}, \texttt{rel-event}, and \texttt{rel-trial} were excluded from this pretraining cohort and are used only for the downstream relational ICL experiments reported in this work. The PluRel configuration instead uses a Griffin checkpoint pretrained on PluRel-generated relational data~\cite{kothapalli2026plurel}. Its downstream evaluation protocol and task sets are otherwise matched to the corresponding Baseline comparison.

\begin{table}[H]
\centering
\caption{\textbf{CTU--RelBench pretraining cohort.}
Databases used to pretrain the CTU--RelBench Griffin checkpoints. The four RelBench databases used for downstream evaluation were held out from this pretraining cohort.}
\label{tab:pretraining_datasets}
\small
\begin{tabular}{lp{0.78\textwidth}}
\toprule
\textbf{Source} & \textbf{Databases} \\
\midrule
CTU &
\texttt{ctu-accidents},
\texttt{ctu-airline},
\texttt{ctu-chess},
\texttt{ctu-dallas},
\texttt{ctu-ergastf1},
\texttt{ctu-financial},
\texttt{ctu-ftp},
\texttt{ctu-geneea},
\texttt{ctu-legalacts},
\texttt{ctu-mondial},
\texttt{ctu-ncaa},
\texttt{ctu-premiereleague},
\texttt{ctu-thrombosis},
\texttt{ctu-tpcc}, and
\texttt{ctu-voc} \\

RelBench &
\texttt{rel-amazon},
\texttt{rel-f1},
\texttt{rel-hm}, and
\texttt{rel-stack} \\
\bottomrule
\end{tabular}
\end{table}

\section{Synthetic Leaker Construction}
\label{app:leakers}

The Full leakage condition contains 20 target-derived columns: 14 on the task target table, three on a directly connected one-hop table, and three on a selected two-hop table. The matched 0-hop, 1-hop, and 2-hop conditions use the same three leaker designs at each distance: a deterministic numerical proxy, a text proxy with a target-informative name and opaque values, and a text proxy with a non-semantic name and target-semantic values. This keeps the proxy family comparable when examining relational placement. The complete construction catalog is given in Table~\ref{tab:leaker_catalog}.

For numerical proxies, rows without an associated support label receive a sentinel value. Noisy proxies preserve the assigned label with probabilities $0.80$, $0.90$, $0.70$, or $0.50$ and otherwise use the alternative class. The partial-coverage proxy is populated on 50\% of the eligible labeled rows. Opaque text values use deterministic class-dependent tokens, whereas semantic text values use task-specific class descriptions processed by the same text-embedding pipeline as original text features. At two hops, labels associated with a destination row are mean-aggregated; numerical proxies store the aggregate directly, while text variants round the aggregate to a class before encoding it. Ground-truth leaker identities are used only to construct the controlled conditions and compute detector metrics.

\begin{table}[t]
\centering
\small
\caption{\textbf{Synthetic leaker catalog.} Name semantics indicates whether the column name identifies its relationship to the target; value semantics indicates whether the stored representation has human-interpretable class meaning.}
\label{tab:leaker_catalog}
\begin{tabular}{clccc}
\toprule
Hop & Leaker type & Encoding & Name sem. & Value sem. \\
\midrule
0 & Perfect deterministic & Numerical & No & No \\
0 & Noisy 80\% & Numerical & No & No \\
0 & Rank/ordinal & Numerical & No & No \\
0 & Jittered numerical proxy & Numerical & No & No \\
0 & Semantic name, opaque values & Text & Yes & No \\
0 & Non-semantic name, semantic values & Text & No & Yes \\
0 & Semantic name and semantic values & Text & Yes & Yes \\
0 & Binned numerical proxy & Numerical & No & No \\
0 & Coarse categorical proxy & Text & No & No \\
0 & Noisy 90\% & Numerical & No & No \\
0 & Noisy 70\% & Numerical & No & No \\
0 & Noisy 50\% & Numerical & No & No \\
0 & Partial-coverage proxy & Numerical & No & No \\
0 & Opaque name and opaque values & Text & No & No \\
1 & Perfect deterministic & Numerical & No & No \\
1 & Semantic name, opaque values & Text & Yes & No \\
1 & Non-semantic name, semantic values & Text & No & Yes \\
2 & Deterministic aggregated proxy & Numerical & No & No \\
2 & Semantic name, opaque values & Text & Yes & No \\
2 & Non-semantic name, semantic values & Text & No & Yes \\
\bottomrule
\end{tabular}
\end{table}

\paragraph{Why nominal hop count does not determine exposure.} Under the temporal sampling rule used in our experiments, an edge whose recorded timestamp is equal to or later than the prediction timestamp is excluded. Consequently, a leaker placed on a neighboring table can affect a target representation only when the corresponding relational path is temporally reachable and selected by the neighborhood sampler. The empirical path audit in Table~\ref{tab:reachability} illustrates the resulting variation across the selected paths. These reachability measurements are diagnostics of the corresponding paths under the evaluation protocol rather than dataset-wide graph statistics.

\begin{table}[t]
\centering
\small
\caption{\textbf{Selected task diagnostics underlying the higher-hop analysis.} Reachability denotes the measured availability of the selected first-hop path under the task timestamps and sampling protocol; values are approximate where reported from the empirical path audit. The class-balance statistics for the Avito tasks refer to the corresponding RelBench test targets. The \texttt{study-outcome} 2-hop entry reports agreement between the rounded aggregate proxy and the queried label.}
\label{tab:reachability}
\begin{tabular}{lll}
\toprule
Task & First-hop exposure & Relevant task/proxy diagnostic \\
\midrule
paper-citation & $\sim$16\% train; $\sim$68\% test & repeated paper labels \\
searchinfo-isuserloggedon & $\sim$0\% in audit & path rarely available \\
searchstream-click & $0\%$ & first edge unavailable \\
user-clicks & $\sim$60\% test & 1.54\% positive in RelBench test \\
user-visits & $\sim$69\% test & 85.06\% positive in RelBench test \\
event-interest tasks & $\sim$4\% & event often occurs later \\
eligibilities-adult & $0\%$ & equal-time first edge \\
eligibilities-child & $0\%$ & equal-time first edge \\
studies-has\_dmc & $0\%$ & first edge unavailable \\
study-outcome & $\sim$91\% & 2-hop proxy-label agreement $\sim$65\% \\
\bottomrule
\end{tabular}
\end{table}

Several task-level patterns are also consistent with simple class-structure diagnostics. The two invariant Avito user-level tasks are strongly imbalanced in the original RelBench test targets: \texttt{user-clicks} contains 1.54\% positive examples and \texttt{user-visits} 85.06\% positive examples~\cite{robinson2024relbench}. Their clean macro-F1 values are close to the corresponding constant-class solutions, providing context for why a support-side perturbation can leave aggregate macro-F1 nearly unchanged. Clean performance for several other tasks is likewise at or near a constant-class solution. These observations should not be interpreted as showing that class imbalance determines the leakage response; rather, class structure can make macro-F1 relatively insensitive to some prediction changes. This is consistent with the broader result that leakage effects depend jointly on task characteristics and effective relational exposure.

\section{Detector Implementation and Additional Results}
\label{app:detection}

\subsection{Ridge-IG scoring}

For IG screening, we first extract the leaky training representations and fit a Ridge classifier ($\alpha=1$) as a differentiable surrogate. We use an internal 80/20 split of the training representations to check the surrogate fit, with stratification when class counts permit, and subsequently evaluate the fitted surrogate on the task validation representations. The Ridge model is used only to define the attribution objective; downstream leakage impact and post-removal performance are evaluated using the corresponding ICL configuration.

IG is computed through the frozen relational encoder on validation batches using 12 midpoint integration steps. The baseline is the zero encoded-feature representation. The sampling seed is reset for each integration point so that the relational neighborhood and internal support context remain fixed along the attribution path. The attribution objective is the margin between the Ridge-predicted class under the full input and its strongest alternative. Signed attributions are integrated before taking their magnitude. For each column, we combine its mean attribution with the mean attribution among its strongest 10\% of occurrences using their geometric mean.

The resulting scores are log-transformed and robustly standardized within source-table--feature-family groups using the median and scaled median absolute deviation; groups containing fewer than ten scored columns use the global statistics. Columns whose robust $z$-score exceeds the empirical 90th percentile for the run are flagged. This q90 rule is an operating point rather than a calibrated statistical test. 

\subsection{MI and LOCO comparison}

We compare IG with MI, which ranks candidate columns by their association with the support labels, and LOCO, which ranks columns according to the downstream change produced by removing one candidate column at a time. Because MI and especially LOCO require substantially more computation, the three-method comparison is restricted to the same Baseline subset: 13 tasks under each of the four leakage conditions, for 52 task--condition pairs. Within each task and condition, IG, MI, and LOCO are evaluated on the same candidate columns and against the same ground-truth leaker identities.

The number of true leakers relative to candidate columns differs between the matched-hop and Full leakage conditions, so the random AUPRC baseline differs across leakage conditions. Within a given task and condition, however, all three detectors use the same candidate set and therefore have the same random baseline.  Table~\ref{tab:detector_summary} reports this mean prevalence together with AUPRC and AUROC.

\begin{table}[h]
\centering
\small
\caption{\textbf{Detector ranking on the common Baseline subset.} All three methods are evaluated on the same task--condition and candidate-column sets within each leakage condition. Random AUPRC is the mean leaker prevalence in these candidate sets.}
\label{tab:detector_summary}
\begin{tabular}{llccc}
\toprule
Method & Condition & AUPRC & AUROC & Random AUPRC \\
\midrule
IG   & 0-hop & 0.596 & 0.906 & 0.052 \\
IG   & 1-hop & 0.128 & 0.568 & 0.052 \\
IG   & 2-hop & 0.051 & 0.282 & 0.052 \\
IG   & Full  & 0.612 & 0.763 & 0.245 \\
\midrule
MI   & 0-hop & 0.419 & 0.689 & 0.052 \\
MI   & 1-hop & 0.120 & 0.574 & 0.052 \\
MI   & 2-hop & 0.122 & 0.585 & 0.052 \\
MI   & Full  & 0.439 & 0.614 & 0.245 \\
\midrule
LOCO & 0-hop & 0.676 & 0.866 & 0.052 \\
LOCO & 1-hop & 0.075 & 0.374 & 0.052 \\
LOCO & 2-hop & 0.061 & 0.413 & 0.052 \\
LOCO & Full  & 0.521 & 0.661 & 0.245 \\
\bottomrule
\end{tabular}
\end{table}

No detector dominates every condition. IG and LOCO provide the strongest 0-hop rankings, while all three methods are substantially weaker in the tested higher-hop settings. IG also remains informative under the Full leakage condition. Because the Full leakage condition has a substantially higher leaker prevalence than the matched-hop conditions, raw AUPRC values should be compared primarily across detector methods within the same condition rather than across leakage conditions. The corresponding AUROC values provide a prevalence-insensitive complementary ranking measure.



\subsection{Threshold and Top-$k$ Verification Sensitivity}

Table~\ref{tab:threshold_sensitivity} reports the thresholded results. Since q90 is a heuristic operating point, we evaluate whether the thresholded detection results depend on this particular choice. We compare q80, q90, and q95 percentile rules with Bonferroni-style $\alpha=0.05$ and BH-style $q=0.05$ score-based operating rules. The latter are included as alternative thresholding heuristics rather than as procedures with formal family-wise error or false-discovery-rate guarantees. Across the percentile rules, decreasing the threshold generally increases recall at the cost of additional false positives, whereas increasing the threshold improves precision while reducing recall. No percentile dominates across conditions: q95 gives the highest 0-hop precision and F1, while q80 gives the highest Full-condition recall and F1. q90 provides an intermediate operating point and yields higher F1 than the Bonferroni-style and BH-style alternatives in both the 0-hop and Full conditions. More importantly, the qualitative condition-level pattern is stable across operating rules: thresholded identification remains substantially stronger for 0-hop and Full leakage than for the tested higher-hop conditions. Thus, the weak higher-hop detection observed at q90 cannot be attributed solely to the choice of the 90th-percentile threshold.

\begin{table}[t]
\centering
\small
\caption{\textbf{IG threshold sensitivity.} Mean thresholded detection metrics across runs under alternative operating rules. Hit rate denotes the fraction of runs in which at least one active leaker is flagged. Bonferroni-style and BH-style thresholds are included as alternative score-based operating rules and are not interpreted as providing formal family-wise error or false-discovery-rate control.}
\label{tab:threshold_sensitivity}
\begin{tabular}{llrrrr}
\toprule
Condition & Rule & Precision & Recall & F1 & Hit rate \\
\midrule
0-hop
& q80 & 0.291 & 0.863 & 0.403 & 0.882 \\
& q90 & 0.357 & 0.716 & 0.430 & 0.882 \\
& q95 & 0.469 & 0.627 & 0.489 & 0.882 \\
& Bonferroni-style ($\alpha=0.05$) & 0.199 & 0.431 & 0.224 & 0.471 \\
& BH-style ($q=0.05$) & 0.195 & 0.490 & 0.244 & 0.529 \\
\midrule
1-hop
& q80 & 0.032 & 0.216 & 0.053 & 0.294 \\
& q90 & 0.009 & 0.039 & 0.015 & 0.118 \\
& q95 & 0.000 & 0.000 & 0.000 & 0.000 \\
& Bonferroni-style ($\alpha=0.05$) & 0.018 & 0.118 & 0.031 & 0.118 \\
& BH-style ($q=0.05$) & 0.017 & 0.118 & 0.030 & 0.118 \\
\midrule
2-hop
& q80 & 0.006 & 0.051 & 0.011 & 0.154 \\
& q90 & 0.000 & 0.000 & 0.000 & 0.000 \\
& q95 & 0.000 & 0.000 & 0.000 & 0.000 \\
& Bonferroni-style ($\alpha=0.05$) & 0.004 & 0.026 & 0.007 & 0.077 \\
& BH-style ($q=0.05$) & 0.004 & 0.026 & 0.007 & 0.077 \\
\midrule
Full
& q80 & 0.461 & 0.398 & 0.411 & 1.000 \\
& q90 & 0.562 & 0.288 & 0.367 & 1.000 \\
& q95 & 0.681 & 0.192 & 0.293 & 0.941 \\
& Bonferroni-style ($\alpha=0.05$) & 0.451 & 0.190 & 0.240 & 0.647 \\
& BH-style ($q=0.05$) & 0.449 & 0.210 & 0.252 & 0.706 \\
\bottomrule
\end{tabular}
\end{table}

We also evaluate a top-$k$ verification protocol (Table \ref{tab:topk_verification}) in which only the $k\in\{1,3,5\}$ highest-ranked columns are inspected. This separates the usefulness of the continuous ranking from the choice of an automatic removal threshold. At 0-hop, inspecting the top five columns identifies at least one active leaker in 88\% of runs and recovers 67\% of active leakers on average. Under Full leakage, the corresponding hit rate is 94\%, although recall is lower because substantially more leakers are active. In contrast, the same small verification budgets recover little or no target-derived information in the tested higher-hop conditions. These results indicate that the ranking can support bounded top-$k$ verification in settings where leakage is strongly exposed, while reinforcing that higher-hop detection remains substantially more difficult.

\begin{table}[t]
\centering
\small
\caption{\textbf{Top-$k$ IG verification.} Mean verification metrics across runs when only the $k\in\{1,3,5\}$ highest-ranked columns are inspected. Precision@$k$ is the fraction of inspected columns that are active leakers, Recall@$k$ is the fraction of active leakers recovered, and Hit rate@$k$ is the fraction of runs in which at least one active leaker appears among the top-$k$ columns.}
\label{tab:topk_verification}
\begin{tabular}{lrrrr}
\toprule
Condition & $k$ & Precision@$k$ & Recall@$k$ & Hit rate@$k$ \\
\midrule
0-hop
& 1 & 0.647 & 0.206 & 0.647 \\
& 3 & 0.510 & 0.520 & 0.882 \\
& 5 & 0.400 & 0.667 & 0.882 \\
\midrule
1-hop
& 1 & 0.000 & 0.000 & 0.000 \\
& 3 & 0.020 & 0.020 & 0.059 \\
& 5 & 0.035 & 0.059 & 0.118 \\
\midrule
2-hop
& 1 & 0.000 & 0.000 & 0.000 \\
& 3 & 0.000 & 0.000 & 0.000 \\
& 5 & 0.000 & 0.000 & 0.000 \\
\midrule
Full
& 1 & 0.471 & 0.034 & 0.471 \\
& 3 & 0.608 & 0.135 & 0.824 \\
& 5 & 0.647 & 0.227 & 0.941 \\
\bottomrule
\end{tabular}
\end{table}

\subsection{Detector-Based Mitigation}
\label{app:mitigation}

Detector-based mitigation removes every flagged column from both support and query schemas and recomputes the relational representations before re-evaluation. This operation is intentionally evaluated without correcting false positives using the known synthetic identities.

The resulting input is generally not identical to the clean condition. False-negative leakers may remain in the support representation, while false-positive original columns are absent from both support and query. When removal leaves a node type without original features, the implementation uses an anonymous zero-valued placeholder representation. Post-removal performance should therefore be interpreted as the outcome of a new feature configuration rather than as an interpolation between clean and leaky data.

We quantify recovery using
\[
G =
\left|
\mathrm{F1}_{\mathrm{leaky}}-
\mathrm{F1}_{\mathrm{clean}}
\right|
-
\left|
\mathrm{F1}_{\mathrm{post}}-
\mathrm{F1}_{\mathrm{clean}}
\right|,
\]
so that $G>0$ denotes movement toward the clean evaluation.

Several runs illustrate why detector quality and safe mitigation are distinct. For TabICL \texttt{user-repeat} at 0-hop, all three active leakers are removed, but ten original columns are also removed; the post-removal degradation therefore cannot be attributed to residual synthetic leakage. In contrast, the PluRel Full \texttt{study-outcome} run removes only six of the twenty active leakers while also removing eight original columns, leaving a substantially different residual-support distribution. A third failure mode occurs when both residual leakage and false-positive removal remain, as in the TabICL Full \texttt{eligibilities-child} run. Representative cases are summarized in Table~\ref{tab:mitigation_cases}. 

\begin{table}[t]
\centering
\small
\caption{\textbf{Representative mitigation failure modes.} Counts refer to active synthetic leakers and original-column false positives removed by the detector.}
\label{tab:mitigation_cases}
\begin{tabular}{llrrrr}
\toprule
Model & Task / condition & TP & FP & FN & Interpretation \\
\midrule
TabICL & user-repeat / 0-hop & 3 & 10 & 0 & original-feature ablation \\
PluRel & study-outcome / Full & 6 & 8 & 14 & residual + ablation \\
TabICL & eligibilities-child / Full & 11 & 3 & 9 & residual + ablation \\
\bottomrule
\end{tabular}
\end{table}

These cases also explain why post-removal macro-F1 occasionally exceeds the clean value. Such an outcome does not demonstrate successful cleaning when target-derived columns remain or original features have been removed. The appropriate mitigation criterion is therefore distance to the clean evaluation together with the detector TP/FP/FN composition, rather than the post-removal score alone.

\section{Individual-Leaker Effects}
\label{app:individual_leakers}

The main experiments evaluate matched hop-specific conditions and the Full 20-column stress test. To determine whether the observed target-table effect is driven by a single synthetic construction, we additionally evaluate each leaker individually using the Baseline Griffin--TabPFN configuration. For each run, the support data contain all original columns and exactly one injected leaker, while the query remains clean. This diagnostic uses a subset of support and query examples per task and is used to examine the distribution of effects across leaker types. We report
\[
\Delta F1 =
F1_{\mathrm{one\text{-}leaker}} - F1_{\mathrm{clean}},
\]
using the clean Baseline from the same diagnostic run.

\begin{figure*}[t]
    \centering
    \includegraphics[width=\textwidth]
    {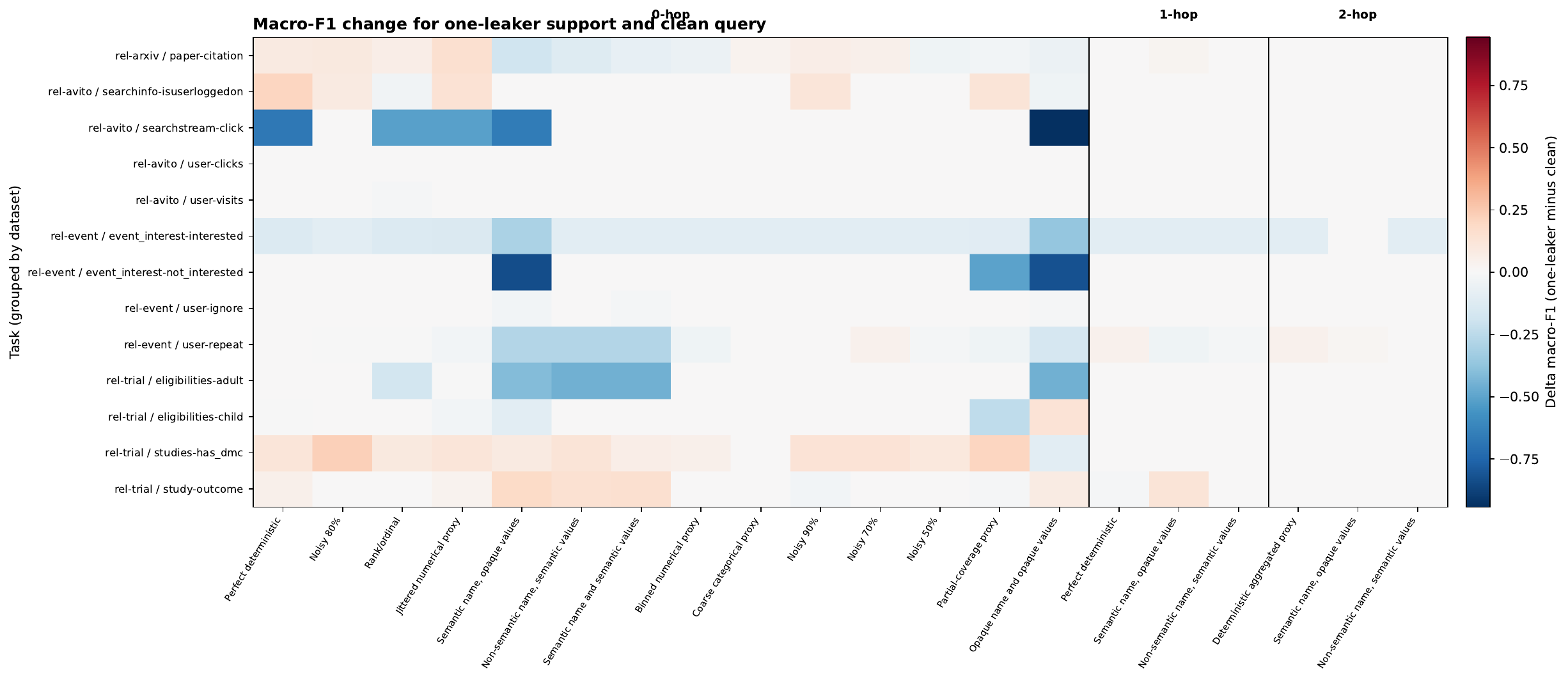}
    \caption{\textbf{Individual-leaker effects under support-only leakage.} Signed change in clean-query macro-F1 when one synthetic leaker at a time is added to the Baseline support context. Rows denote the 13 RelBench tasks and columns denote the 20 synthetic  constructions, grouped by relational placement. Negative and positive values indicate degradation and improvement, respectively, relative to the corresponding clean
    Baseline evaluation.}
    \label{fig:individual_leakers}
\end{figure*}

Figure~\ref{fig:individual_leakers} shows that the target-table effect is not specific to an exact label copy (perfect deterministic) or to one representation type. Across the 182 task--leaker conditions at 0-hop, mean absolute $|\Delta F1|$ is 0.083, compared with 0.014 and 0.007 across the 39 conditions at one and two hops,
respectively. All 14 target-table constructions produce $|\Delta F1|>0.05$ on at least one task. Thus, the aggregate 0-hop effect cannot be attributed to one specially chosen synthetic leaker type.

The magnitude and direction of the individual effects remain strongly task dependent. Several distinct 0-hop constructions produce substantial changes on \texttt{searchstream-click}, the event-interest tasks, and \texttt{eligibilities-adult}, whereas \texttt{user-clicks} is unchanged across all individual constructions and \texttt{user-visits} is nearly invariant. Positive as well as negative changes also occur, consistent with the main analysis that support-set contamination should be measured by deviation from the clean evaluation rather than by assuming a fixed direction of effect.

The individual higher-hop effects are substantially sparser: only 4/39 one-hop and 2/39 two-hop conditions have $|\Delta F1|>0.05$, compared with 65/182 at 0-hop. They are not uniformly zero, however. This is consistent with the aggregate hop-specific results in  Figure~\ref{fig:figure2_model_task_hop_leaker_type} and with the interpretation that nominal relational distance alone does not determine leakage severity. Whether a higher-hop target-derived feature affects the representation also depends on temporal reachability, sampled exposure, and aggregation fidelity, as detailed in Appendix~\ref{app:leakers}.

Finally, the one-leaker results do not support a universal ordering by noise level, semantic transparency, or representation type. The same construction can have a large effect on one task and little or oppositely directed effect on another. We therefore use the leaker families as a controlled stress-test suite rather than claiming that any individual leaker family is universally more severe. Because this diagnostic uses the Baseline configuration only, it is not used to infer model-specific leaker-family preferences. 

\section{Complete Task-Level Effects and Ranking Reversals}
\label{app:full_results}

Table~\ref{tab:all_effects} reports the complete task-level values underlying Figure~2A. We report the clean macro-F1 together with the signed change under each support-only leakage condition. Positive and negative values are retained because the primary quantity of interest is deviation from the valid clean evaluation rather than performance degradation alone.

\begin{table*}[t]
\centering
\scriptsize
\caption{\textbf{Complete task-level support-only leakage effects.} Each $\Delta$F1 is leaky-support/clean-query macro-F1 minus the corresponding clean macro-F1.}
\label{tab:all_effects}
\resizebox{\textwidth}{!}{%
\begin{tabular}{lllrrrrr}
\toprule
Model & Dataset & Task & Clean F1 & 0-hop delta F1 & 1-hop delta F1 & 2-hop delta F1 & Full delta F1 \\
\midrule
Baseline & rel-arxiv & paper-citation & 0.6719 & -0.0729 & 0.0000 & -0.0103 & -0.1828 \\
Baseline & rel-avito & searchinfo-isuserloggedon & 0.4174 & 0.0146 & -0.0070 & -0.0070 & -0.1711 \\
Baseline & rel-avito & searchstream-click & 1.0000 & -0.6452 & 0.0000 & 0.0000 & -0.7987 \\
Baseline & rel-avito & user-clicks & 0.4960 & 0.0000 & 0.0000 & 0.0000 & 0.0000 \\
Baseline & rel-avito & user-visits & 0.4698 & 0.0000 & 0.0000 & 0.0000 & 0.0000 \\
Baseline & rel-event & event\_interest-interested & 0.4370 & -0.2209 & 0.0000 & 0.0000 & -0.0445 \\
Baseline & rel-event & event\_interest-not\_interested & 0.4976 & 0.0000 & 0.0000 & 0.0000 & -0.1229 \\
Baseline & rel-event & user-ignore & 0.6812 & -0.1067 & 0.0022 & 0.0022 & -0.2047 \\
Baseline & rel-event & user-repeat & 0.6324 & -0.1202 & 0.0449 & 0.0566 & -0.1279 \\
Baseline & rel-trial & eligibilities-adult & 0.4829 & -0.2951 & 0.0000 & 0.0000 & -0.4210 \\
Baseline & rel-trial & eligibilities-child & 0.4583 & -0.0065 & 0.0000 & 0.0000 & 0.0000 \\
Baseline & rel-trial & studies-has\_dmc & 0.5560 & -0.1381 & 0.0015 & 0.0015 & -0.2432 \\
Baseline & rel-trial & study-outcome & 0.5139 & -0.1439 & 0.0262 & -0.0513 & -0.0044 \\
MP-8 & rel-arxiv & paper-citation & 0.6072 & -0.1447 & 0.0401 & 0.0000 & -0.1405 \\
MP-8 & rel-avito & searchinfo-isuserloggedon & 0.4833 & -0.0354 & 0.0348 & 0.0348 & -0.0232 \\
MP-8 & rel-avito & searchstream-click & 1.0000 & -0.9174 & 0.0000 & 0.0000 & -0.7143 \\
MP-8 & rel-avito & user-clicks & 0.4960 & 0.0000 & 0.0000 & 0.0000 & 0.0000 \\
MP-8 & rel-avito & user-visits & 0.4698 & 0.0000 & 0.0000 & 0.0000 & 0.0000 \\
MP-8 & rel-event & event\_interest-interested & 0.4370 & 0.0249 & 0.0000 & 0.0000 & -0.2507 \\
MP-8 & rel-event & event\_interest-not\_interested & 0.4976 & 0.0000 & 0.0000 & 0.0000 & -0.0155 \\
MP-8 & rel-event & user-ignore & 0.5814 & -0.1050 & 0.0108 & -0.0446 & -0.1050 \\
MP-8 & rel-event & user-repeat & 0.6991 & -0.2072 & -0.0411 & -0.0285 & -0.1741 \\
MP-8 & rel-trial & eligibilities-adult & 0.4829 & -0.3202 & 0.0000 & 0.0000 & -0.2694 \\
MP-8 & rel-trial & eligibilities-child & 0.4583 & 0.0480 & 0.0000 & 0.0000 & -0.3248 \\
MP-8 & rel-trial & studies-has\_dmc & 0.5076 & -0.0200 & 0.0012 & -0.0072 & 0.0141 \\
MP-8 & rel-trial & study-outcome & 0.3723 & 0.1153 & 0.0398 & -0.0093 & 0.0732 \\
PluRel & rel-arxiv & paper-citation & 0.6589 & -0.1606 & -0.0442 & 0.0030 & -0.2648 \\
PluRel & rel-avito & searchinfo-isuserloggedon & 0.4118 & 0.0944 & 0.0000 & 0.0000 & 0.0832 \\
PluRel & rel-avito & searchstream-click & 1.0000 & 0.0000 & 0.0000 & 0.0000 & -0.9009 \\
PluRel & rel-avito & user-clicks & 0.4960 & 0.0000 & 0.0000 & 0.0000 & 0.0000 \\
PluRel & rel-avito & user-visits & 0.4698 & 0.0000 & 0.0000 & 0.0000 & 0.0000 \\
PluRel & rel-event & event\_interest-interested & 0.4370 & 0.0196 & 0.0000 & 0.0000 & -0.2541 \\
PluRel & rel-event & event\_interest-not\_interested & 0.4976 & -0.0155 & 0.0000 & 0.0000 & -0.4882 \\
PluRel & rel-event & user-ignore & 0.6412 & -0.1435 & 0.0085 & 0.0055 & -0.1648 \\
PluRel & rel-event & user-repeat & 0.6481 & -0.2501 & -0.0233 & -0.0070 & -0.1943 \\
PluRel & rel-trial & eligibilities-adult & 0.4829 & -0.4210 & 0.0000 & 0.0000 & -0.4210 \\
PluRel & rel-trial & eligibilities-child & 0.4583 & 0.0000 & 0.0000 & 0.0000 & 0.0000 \\
PluRel & rel-trial & studies-has\_dmc & 0.4862 & -0.0448 & -0.0071 & -0.0071 & -0.0448 \\
PluRel & rel-trial & study-outcome & 0.4549 & 0.0168 & -0.0918 & -0.0019 & -0.1154 \\
TabICL & rel-arxiv & paper-citation & 0.6417 & -0.2133 & -0.0806 & -0.0138 & -0.2693 \\
TabICL & rel-avito & searchinfo-isuserloggedon & 0.5986 & -0.1486 & -0.1164 & -0.1205 & -0.0989 \\
TabICL & rel-avito & searchstream-click & 1.0000 & -0.9921 & 0.0000 & 0.0000 & -0.9804 \\
TabICL & rel-avito & user-clicks & 0.4960 & 0.0000 & 0.0000 & 0.0000 & 0.0000 \\
TabICL & rel-avito & user-visits & 0.4698 & 0.0000 & 0.0000 & 0.0000 & 0.0000 \\
TabICL & rel-event & event\_interest-interested & 0.4370 & 0.0196 & 0.0000 & 0.0000 & 0.0000 \\
TabICL & rel-event & event\_interest-not\_interested & 0.4976 & 0.0000 & 0.0000 & 0.0000 & 0.0000 \\
TabICL & rel-event & user-ignore & 0.6286 & -0.1521 & -0.0412 & -0.0951 & -0.1521 \\
TabICL & rel-event & user-repeat & 0.6459 & 0.0061 & 0.0484 & 0.0488 & 0.0080 \\
TabICL & rel-trial & eligibilities-adult & 0.4829 & -0.3857 & 0.0000 & 0.0000 & -0.4210 \\
TabICL & rel-trial & eligibilities-child & 0.4583 & 0.0337 & 0.0000 & 0.0000 & 0.0505 \\
TabICL & rel-trial & studies-has\_dmc & 0.5088 & -0.0596 & -0.0390 & -0.0390 & -0.0596 \\
TabICL & rel-trial & study-outcome & 0.3985 & 0.1409 & 0.0791 & -0.0303 & 0.0811 \\
Bimodal & rel-arxiv & paper-citation & 0.6753 & -0.3064 & 0.0105 & 0.0146 & -0.3138 \\
Bimodal & rel-avito & searchinfo-isuserloggedon & 0.4118 & 0.0854 & 0.0000 & 0.0000 & 0.0495 \\
Bimodal & rel-avito & searchstream-click & 1.0000 & -0.8696 & 0.0000 & 0.0000 & -0.9785 \\
Bimodal & rel-avito & user-clicks & 0.4960 & 0.0000 & 0.0000 & 0.0000 & 0.0000 \\
Bimodal & rel-avito & user-visits & 0.4698 & 0.0000 & 0.0000 & 0.0000 & 0.0000 \\
Bimodal & rel-event & event\_interest-interested & 0.4370 & 0.0196 & 0.0000 & 0.0000 & 0.0196 \\
Bimodal & rel-event & event\_interest-not\_interested & 0.4976 & -0.0155 & 0.0000 & 0.0000 & -0.0845 \\
Bimodal & rel-event & user-ignore & 0.6663 & -0.1898 & -0.0321 & -0.0042 & -0.1898 \\
Bimodal & rel-event & user-repeat & 0.6150 & 0.0104 & 0.0548 & 0.0026 & -0.0138 \\
Bimodal & rel-trial & eligibilities-adult & 0.4829 & 0.0000 & 0.0000 & 0.0000 & -0.4069 \\
Bimodal & rel-trial & eligibilities-child & 0.4583 & -0.3248 & 0.0000 & 0.0000 & -0.3248 \\
Bimodal & rel-trial & studies-has\_dmc & 0.5124 & -0.0488 & -0.0262 & -0.0262 & -0.2794 \\
Bimodal & rel-trial & study-outcome & 0.4374 & 0.1489 & -0.0292 & 0.0260 & -0.0000 \\
\bottomrule
\end{tabular}

}
\end{table*}

\paragraph{Strict ranking reversals.}
We count a strict reversal when the sign of a variant's performance difference from the Baseline changes between clean and support-only leakage evaluation. Comparisons with an absolute difference at or below $10^{-6}$ under either condition are treated as ties and excluded from the denominator. Table~\ref{tab:flip_summary} reports the resulting counts.

\begin{table}[t]
\centering
\small
\caption{\textbf{Strict model-ranking reversals.} Entries report reversals/non-tied comparisons.}
\label{tab:flip_summary}
\begin{tabular}{lcccc}
\toprule
Variant & 0-hop & 1-hop & 2-hop & Full \\
\midrule
MP-8    & 3/6 & 1/6 & 1/6 & 1/5 \\
PluRel  & 4/6 & 2/6 & 3/6 & 3/5 \\
TabICL  & 2/6 & 0/6 & 0/6 & 1/5 \\
Bimodal & 5/6 & 1/6 & 2/6 & 3/5 \\
\bottomrule
\end{tabular}
\end{table}

For completeness, the task-level comparisons that contribute to these counts are reported in Table~\ref{tab:individual_flips}.

\begin{table*}[t]
\centering
\scriptsize
\caption{\textbf{Task-level strict ranking reversals.} The two difference columns are variant minus Baseline macro-F1 under the corresponding evaluation condition.}
\label{tab:individual_flips}
\resizebox{\textwidth}{!}{%
\begin{tabular}{llllrrr}
\toprule
Variant & Dataset & Task & Scenario & Variant-Baseline clean & Variant-Baseline leaky & Strict flip \\
\midrule
Bimodal & rel-arxiv & paper-citation & 0-hop & 0.0034 & -0.2301 & True \\
PluRel & rel-arxiv & paper-citation & 2-hop & -0.0130 & 0.0003 & True \\
Bimodal & rel-arxiv & paper-citation & Full & 0.0034 & -0.1277 & True \\
Bimodal & rel-avito & searchinfo-isuserloggedon & 0-hop & -0.0056 & 0.0651 & True \\
PluRel & rel-avito & searchinfo-isuserloggedon & 0-hop & -0.0056 & 0.0742 & True \\
Bimodal & rel-avito & searchinfo-isuserloggedon & 1-hop & -0.0056 & 0.0014 & True \\
PluRel & rel-avito & searchinfo-isuserloggedon & 1-hop & -0.0056 & 0.0014 & True \\
Bimodal & rel-avito & searchinfo-isuserloggedon & 2-hop & -0.0056 & 0.0014 & True \\
PluRel & rel-avito & searchinfo-isuserloggedon & 2-hop & -0.0056 & 0.0014 & True \\
Bimodal & rel-avito & searchinfo-isuserloggedon & Full & -0.0056 & 0.2150 & True \\
PluRel & rel-avito & searchinfo-isuserloggedon & Full & -0.0056 & 0.2487 & True \\
Bimodal & rel-event & user-repeat & 0-hop & -0.0173 & 0.1133 & True \\
MP-8 & rel-event & user-repeat & 0-hop & 0.0667 & -0.0202 & True \\
PluRel & rel-event & user-repeat & 0-hop & 0.0157 & -0.1142 & True \\
MP-8 & rel-event & user-repeat & 1-hop & 0.0667 & -0.0193 & True \\
PluRel & rel-event & user-repeat & 1-hop & 0.0157 & -0.0525 & True \\
MP-8 & rel-event & user-repeat & 2-hop & 0.0667 & -0.0184 & True \\
PluRel & rel-event & user-repeat & 2-hop & 0.0157 & -0.0479 & True \\
Bimodal & rel-event & user-repeat & Full & -0.0173 & 0.0967 & True \\
PluRel & rel-event & user-repeat & Full & 0.0157 & -0.0508 & True \\
Bimodal & rel-trial & studies-has\_dmc & 0-hop & -0.0435 & 0.0458 & True \\
MP-8 & rel-trial & studies-has\_dmc & 0-hop & -0.0484 & 0.0697 & True \\
PluRel & rel-trial & studies-has\_dmc & 0-hop & -0.0698 & 0.0235 & True \\
TabICL & rel-trial & studies-has\_dmc & 0-hop & -0.0472 & 0.0313 & True \\
MP-8 & rel-trial & studies-has\_dmc & Full & -0.0484 & 0.2089 & True \\
PluRel & rel-trial & studies-has\_dmc & Full & -0.0698 & 0.1286 & True \\
TabICL & rel-trial & studies-has\_dmc & Full & -0.0472 & 0.1363 & True \\
Bimodal & rel-trial & study-outcome & 0-hop & -0.0765 & 0.2163 & True \\
MP-8 & rel-trial & study-outcome & 0-hop & -0.1415 & 0.1177 & True \\
PluRel & rel-trial & study-outcome & 0-hop & -0.0590 & 0.1017 & True \\
TabICL & rel-trial & study-outcome & 0-hop & -0.1154 & 0.1695 & True \\
Bimodal & rel-trial & study-outcome & 2-hop & -0.0765 & 0.0007 & True \\
\bottomrule
\end{tabular}

}
\end{table*}

The higher-hop reversals do not contradict the near-zero aggregate $|\Delta\mathrm{F1}|$ values. Aggregate summaries measure the typical magnitude of a perturbation, whereas model selection depends on whether the task-specific leakage shift exceeds the clean performance margin between two models. Consequently, small effects can leave the aggregate median almost unchanged while still changing which model is preferred.

\paragraph{Ranking-margin sensitivity.}
The primary analysis treats variant--Baseline differences with absolute magnitude at or below $10^{-6}$ as ties. To assess whether the ranking-reversal result is driven by comparisons with very small model-performance margins, we repeat the analysis while requiring both the clean and support-only variant--Baseline differences to exceed a minimum absolute margin $\epsilon$. Of the 32 strict reversals in the primary analysis, 24 remain at $\epsilon=0.005$, 20 remain at $\epsilon=0.01$, and 12 remain at $\epsilon=0.02$. Thus, although some reversals involve small performance margins, the model-ranking effect is not restricted to near-tied comparisons (Table \ref{tab:ranking_margin_sensitivity}).

\begin{table}[h]
\centering
\small
\caption{\textbf{Ranking-reversal margin sensitivity.} Number of strict reversals retained when both the clean and support-only variant--Baseline performance margins must exceed $\epsilon$ in absolute value.}
\label{tab:ranking_margin_sensitivity}
\begin{tabular}{lr}
\toprule
Minimum margin $\epsilon$ & Reversals retained \\
\midrule
$0$ & 32 \\
$0.005$ & 24 \\
$0.01$ & 20 \\
$0.02$ & 12 \\
\bottomrule
\end{tabular}
\end{table}

\section{\texttt{rel-salt} Native-Schema Target-Field Leakage Case Study}
\label{app:salt}

We additionally evaluate four multiclass \texttt{rel-salt} tasks using the Baseline configuration (Table \ref{tab:salt}). No synthetic columns are introduced. For each task, we compare a reference feature configuration in which the other seven SALT target-designated fields are excluded with a target-field-included configuration in which these fields are available in the relational context. We then apply the same detector-based removal procedure. Because task difficulty varies substantially across these tasks, we interpret this case study primarily through paired within-task changes relative to the target-field-excluded reference rather than through absolute macro-F1 alone.

\begin{table}[H]
\centering
\small
\caption{\textbf{\texttt{rel-salt} target-field case study.} $\Delta$ Included and $\Delta$ Post report macro-F1 change relative to this task-specific reference when the other seven target-designated fields are included and after detector-based removal, respectively. TP counts detector-flagged target-designated candidate fields, while FP counts flagged other original features; these labels are defined relative to the candidate set used in this case study and do not imply that every target-designated field is a confirmed deployment-time leaker.}
\label{tab:salt}
\begin{tabular}{lrrrrr}
\toprule
Task & $\Delta$ Included & $\Delta$ Post & TP & FP \\
\midrule
item-incoterms  & -0.008 & -0.008 & 2 & 1 \\
item-plant      & -0.033 & -0.010 & 1 & 2 \\
sales-incoterms & -0.235 & -0.252 & 2 & 1 \\
sales-office     & 0.000 & 0.000 & 3 & 0 \\
\midrule
Mean             & -0.069 & -0.068 & -- & -- \\
\bottomrule
\end{tabular}
\end{table}

The paired effects are heterogeneous across tasks. Including the additional target-designated fields produces little or no change for \texttt{item-incoterms} and \texttt{sales-office}, but larger shifts for \texttt{item-plant} ($\Delta\mathrm{F1}=-0.033$) and especially \texttt{sales-incoterms} ($\Delta\mathrm{F1}=-0.235$). Averaged across the four tasks, target-field inclusion changes macro-F1 by $-0.069$ relative to the corresponding reference configurations. Detector-based removal does not consistently restore these reference results: for \texttt{item-plant}, the post-removal difference decreases from $-0.033$ to $-0.010$, whereas for \texttt{sales-incoterms} it increases in magnitude from $-0.235$ to $-0.252$.

Thus, the native-schema case study exhibits the same qualitative evaluation concern as the controlled experiments: additional target-designated fields can materially alter measured performance on some tasks, while automatic removal need not recover the corresponding reference feature configuration. We treat this experiment as a native-schema target-field case study rather than as evidence that every additional SALT target-designated field is a confirmed deployment-time leaker. Establishing the latter requires field- and task-specific prediction-time availability semantics.

\section{Limitations}
\label{app:limitations}

The synthetic experiments are controlled stress tests of a specific support/query availability mismatch and do not enumerate every form of leakage that can occur in relational databases. The primary benchmark contains 13 binary RelBench tasks and five relational ICL configurations; the rel-salt analysis contains four multiclass tasks and is evaluated with the Baseline configuration only.

The reported experiments use a fixed evaluation/subsampling configuration rather than estimating variance across repeated support-set samples. Accordingly, the task- and model-dependent effects reported here should not be interpreted as estimates of population-level leakage frequency or support-sampling variance.

The higher-hop analysis depends on the selected relational paths, temporal reachability, neighborhood sampling, and aggregation used in these experiments. The measured reachability statistics support the aggregate explanation of weak higher-hop exposure, but per-query sampled-subgraph traces would be required to assign every individual prediction change causally to a specific path.

Ridge-IG attributes a surrogate objective over relational representations rather than the downstream TabPFN or TabICLv2 prediction itself. The q90 threshold and the alternative score-based operating rules evaluated in Appendix~\ref{app:detection} are heuristic decision rules rather than calibrated hypothesis tests. The detector results therefore separate continuous ranking quality from thresholded identification and from the downstream consequences of feature removal. The current Ridge-IG detector should accordingly be viewed as an initial attribution-based screening approach rather than a complete solution to relational leakage detection. Future work will investigate graph- and relation-aware attribution methods that explicitly account for relational paths, neighborhoods, and higher-hop information flow, and assess whether they improve detection when target-derived information is only indirectly exposed to the encoder.

Finally, the rel-salt experiment evaluates native-schema leakage candidates but does not independently establish the prediction-time availability of every candidate field. It is therefore reported separately from the controlled RelBench benchmark.

\newpage

\end{document}